\documentclass[conference,a4paper]{IEEEtran}

\usepackage[utf8]{inputenc}
\usepackage[T1]{fontenc}
\usepackage{amsmath,amssymb}
\usepackage{booktabs}
\usepackage{multirow}
\usepackage{graphicx}
\usepackage{xcolor}
\usepackage{colortbl}
\usepackage{caption}
\usepackage{subcaption}
\usepackage{microtype}
\usepackage{gensymb}
\usepackage{cite}
\usepackage{tikz}
\usetikzlibrary{arrows.meta}
\usepackage{hyperref}
\hypersetup{
  hidelinks,
  pdftitle={Client-Conditional Federated Learning via Local Training Data Statistics},
  pdfauthor={Rickard Brännvall},
  pdfsubject={Federated learning, personalization, data heterogeneity},
  pdfkeywords={federated learning, personalization, data heterogeneity, PCA statistics}
}

\definecolor{bestcolor}{RGB}{215,94,0}
\definecolor{oraclecolor}{RGB}{150,150,150}

\newcommand{\best}[1]{\textbf{#1}}

\begin{document}

\title{Client-Conditional Federated Learning via Local Training Data Statistics}

\author{\IEEEauthorblockN{Rickard Br\"annvall}
\IEEEauthorblockA{RISE Research Institutes of Sweden\\
Email: rickard.brannvall@ri.se}}

\onecolumn
\thispagestyle{empty}

\begin{center}
{\LARGE\bfseries Client-Conditional Federated Learning\\via Local Training Data Statistics \par}
\vskip 1em
{\large Rickard Br\"annvall \par}
\vskip 0.2em
RISE Research Institutes of Sweden \par
\vskip 0.1em
\texttt{rickard.brannvall@ri.se} \par
\vskip 1em
\emph{Extended version of the FLICS 2026 paper~\cite{brannvall2026flics_camera},\\
with full experimental tables and figures provided as appendices.}
\end{center}

\vskip 0.8em
\noindent\textbf{Abstract.}\quad
Federated learning (FL) under data heterogeneity remains challenging: existing methods either ignore client differences (FedAvg), require costly cluster discovery (IFCA), or maintain per-client models (Ditto). All degrade when data is sparse or heterogeneity is multi-dimensional. We propose conditioning a single global model on locally-computed PCA statistics of each client's training data, requiring no additional communication or disclosure of client information. Evaluating across 97~configurations spanning four heterogeneity types (label shift, covariate shift, concept shift, and combined heterogeneity), four datasets (MNIST, Fashion-MNIST, CIFAR-10, CIFAR-100), and seven FL baseline methods, we find that our method matches the Oracle baseline---which knows true cluster assignments---across all settings, surpasses it by 1--6\% on combined heterogeneity where continuous statistics are richer than discrete cluster identifiers, and is uniquely sparsity-robust among all tested methods.

\vskip 0.8em
\setcounter{tocdepth}{1}
{\small\tableofcontents}

\newpage

\twocolumn
\pagestyle{plain}        
\setcounter{page}{1}     
\maketitle
\thispagestyle{plain}    

\begin{abstract}
Federated learning (FL) under data heterogeneity remains challenging: existing methods either ignore client differences (FedAvg), require costly cluster discovery (IFCA), or maintain per-client models (Ditto).
All degrade when data is sparse or heterogeneity is multi-dimensional.
We propose conditioning a single global model on locally-computed PCA statistics of each client's training data, requiring no additional communication or disclosure of client information.
Evaluating across 97~configurations spanning four heterogeneity types (label shift, covariate shift, concept shift, and combined heterogeneity), four datasets (MNIST, Fashion-MNIST, CIFAR-10, CIFAR-100), and seven FL baseline methods, we find that our method matches the Oracle baseline---which knows true cluster assignments---across all settings, surpasses it by 1--6\% on combined heterogeneity where continuous statistics are richer than discrete cluster identifiers, and is uniquely sparsity-robust among all tested methods.
\end{abstract}

\begin{IEEEkeywords}
Federated learning, personalization, data heterogeneity, PCA statistics
\end{IEEEkeywords}

\section{Introduction}
\label{sec:intro}

Federated learning (FL) enables collaborative model training across distributed clients without sharing raw data~\cite{mcmahan2017fedavg, kairouz2021advances, arbaoui2024fedsurvey}.
A core promise of FL is that collaboration improves upon what each client could achieve alone: by pooling gradient information, the shared model can generalize better than any single client's local model.
However, this promise breaks down under \emph{data heterogeneity}---the realistic situation where different clients have different data distributions~\cite{zhao2018noniid}.

Heterogeneity manifests in multiple forms.
Under \emph{label shift}~(E1), different clients observe different subsets of classes, so averaging their gradients toward conflicting class boundaries destroys information.
Under \emph{covariate shift}~(E2), clients observe different feature distributions even for the same task, causing the global model to average over incompatible input statistics.
Under \emph{concept shift}~(E3), clients may apply different classification rules to the same inputs, making their objectives fundamentally incompatible.
When these forms combine~(E4), the challenge is compounded.

The standard federated averaging algorithm (FedAvg)~\cite{mcmahan2017fedavg} produces a single global model that compromises across all local objectives.
Under severe heterogeneity, this compromise can be catastrophic: our experiments show FedAvg accuracy dropping from 73.5\% to 17.2\% on CIFAR-10 as the number of distinct client clusters increases from 2 to 10.

Existing approaches to heterogeneous FL fall into three categories, each with significant limitations:
\begin{itemize}
\item \textbf{Clustered FL} methods (IFCA~\cite{ghosh2020ifca}, CFL~\cite{sattler2020cfl}) attempt to discover groups of similar clients and train separate models per group. They require iterative cluster estimation, which is unreliable under data sparsity, disclose additional information, and assume that a small number of discrete clusters fully capture the heterogeneity structure.
\item \textbf{Personalized FL} methods (Ditto~\cite{li2021ditto}, Per-FedAvg~\cite{fallah2020perfedavg}) maintain per-client model components, increasing communication and storage with the number of clients.
\item \textbf{Decentralized FL} methods (DAC~\cite{zec2022dac}) use peer-to-peer communication weighted by model similarity, but still require iterative discovery of the collaboration structure.
\end{itemize}

All three categories share a common weakness: they \emph{discover} client similarity from model behavior---gradient directions, loss values, or weight distances---which becomes unreliable when individual clients have little data (sparsity) or when heterogeneity has multiple simultaneous dimensions.
Furthermore, the discovery process itself---exchanging cluster assignments, model similarity scores, or pairwise loss evaluations---can disclose sensitive information about clients' data distributions beyond what standard federated averaging reveals.

\begin{table}
\centering
\caption{Summary of representative results across all experiments E1--E4. \best{Bold} indicates best non-Oracle method. Conditional consistently matches or exceeds Oracle.}
\label{tab:summary}
\setlength{\tabcolsep}{3pt}
\footnotesize
\begin{tabular}{@{}llcccc@{}}
\toprule
Exp. & Config & FedAvg & Oracle & \best{Cond} & Best Other \\
\midrule
E1 & CIFAR-10 $K{=}5$ & .325 & .927 & \best{.929} & .906 (IFCA) \\
E2a & CIFAR-10 $K{=}3$ & .943 & .979 & \best{.981} & .972 (DAC) \\
E2b & FMNIST rot. $K{=}4$ & .789 & .900 & \best{.895} & .868 (DAC) \\
E3a & FMNIST sem. Rich & .671 & .950 & \best{.950} & .943 (IFCA) \\
E3b & CIFAR-10 perm. $K{=}2$ & .408 & .795 & \best{.808} & .773 (DAC) \\
E4a & MNIST+FM Rich & .939 & .957 & \best{.955} & .953 (IFCA) \\
E4b & CIFAR-10 $C{=}3$ & .743 & .914 & \best{.935} & .880 (Ditto) \\
\bottomrule
\end{tabular}
\end{table}

In this paper, we take a fundamentally different approach.
Instead of discovering client relationships, we \emph{characterize} each client's data distribution directly.
Each client computes a compact vector of PCA eigenvalues on its local data, capturing the principal modes of variation in its joint feature-label distribution.
A single global model is then conditioned on this statistics vector via concatenation before the fully-connected layers, enabling it to adapt its predictions to each client's distribution.
No separate models are maintained, no cluster structure is discovered, and no client statistics are communicated.

Our contributions are:
\begin{enumerate}
\item A method that conditions a single federated model on locally-computed training data PCA statistics, requiring \textbf{no additional communication or disclosure of client information} beyond standard FL (Sec.~\ref{sec:method}).
\item A comprehensive evaluation across \textbf{97 configurations}: 4~heterogeneity types (label shift, covariate shift, concept shift, combined), 4~datasets, 7~baselines, and sparsity ranging from 6{,}000 to 200 samples per client (Sec.~\ref{sec:experiments}).
\item Evidence that \textbf{continuous distribution statistics outperform discrete cluster knowledge}: our method surpasses the Oracle baseline (which knows true cluster assignments) by 1--6\% on combined heterogeneity, because continuous statistics capture variation that a discrete cluster ID cannot (Sec.~\ref{sec:claim2}).
\item Demonstration of \textbf{unique sparsity robustness}: our method maintains near-constant accuracy as client data decreases 20-fold from 6{,}000 to 200 samples, while all other tested methods degrade by 6--85\% (Sec.~\ref{sec:claim3}).
\end{enumerate}

\section{Related Work}
\label{sec:related}

\begin{figure}[t]
\centering
\begin{tikzpicture}[
  >=Stealth,
  cbox/.style={draw=gray!70, thick, rounded corners=2pt, minimum height=7mm, font=\footnotesize, inner sep=3pt, fill=green!10},
  gbox/.style={draw=gray!70, thick, rounded corners=2pt, minimum height=7mm, font=\footnotesize\bfseries, inner sep=3pt, fill=orange!15},
  stagelbl/.style={font=\footnotesize\bfseries, anchor=west},
  arr/.style={->, thick, gray!60},
  sep/.style={dashed, gray!40},
]
\def\ya{2.8}   
\def\yd{1.8}   
\def\yb{0.8}   
\def\yc{-0.6}  
\def\xr{7.8}   
\def\xd{1.4}   

\node[cbox] (data) at (\xd, \yd) {$(\mathbf{x}, y) \in \mathcal{D}_i^{\text{train}}$};

\node[stagelbl] at (-0.5, \ya) {Prepare:};
\node[cbox] (zi) at (3.8, \ya) {$\bigl[\phi(\mathbf{x}) \,\|\, \text{onehot}(y)\bigr]$};
\node[cbox, font=\footnotesize\bfseries] (pca) at (6.0, \ya) {PCA};
\node[cbox] (si) at (\xr, \ya) {$\mathbf{s}_i$};
\draw[arr] (data.north) |- (zi.west);
\draw[arr] (zi) -- (pca);
\draw[arr] (pca) -- (si);

\draw[sep] (-0.5, 1.3) -- (\xr+0.6, 1.3);
\node[stagelbl] at (-0.5, \yb) {Train:};
\node[cbox] (loss) at (3.8, \yb) {$\nabla_{\boldsymbol{\theta}}\,\mathcal{L}\bigl(y,\, f(\mathbf{x};\boldsymbol{\theta},\mathbf{s}_i)\bigr)$};
\node[gbox] (fl) at (6.0, \yb) {FL};
\node[gbox] (theta) at (\xr, \yb) {$\boldsymbol{\theta}$};
\draw[arr] (data.south) |- (loss.west);
\draw[arr] (loss) -- (fl);
\draw[arr] (fl) -- (theta);

\draw[sep] (-0.5, 0.1) -- (\xr+0.6, 0.1);
\node[stagelbl] at (-0.5, \yc) {Infer:};
\node[cbox] (xtest) at (\xd, \yc) {$\mathbf{x}
\in \mathcal{D}_i^{\text{test}}$};
\node[cbox] (finfer) at (3.8, \yc) {$f(\mathbf{x};\boldsymbol{\theta},\mathbf{s}_i)$};
\node[cbox] (yhat) at (\xr, \yc) {$\hat{y}$};
\draw[arr] (xtest) -- (finfer);
\draw[arr] (finfer) -- (yhat);
\end{tikzpicture}
\caption{Client-conditional pipeline for client~$i$. Green boxes are client-local; orange boxes involve the federation. \textbf{Prepare}: PCA eigenvalues~$\mathbf{s}_i$ are computed once from the training data. \textbf{Train}: model updates are computed locally and aggregated via federated learning to produce the shared model~$\boldsymbol{\theta}$. \textbf{Infer}: the client uses the shared model $\boldsymbol{\theta}$ and its own $\mathbf{s}_i$ for predictions.}
\label{fig:pipeline}
\end{figure}

\textbf{Clustered federated learning.}
IFCA~\cite{ghosh2020ifca} maintains $K$ cluster models and assigns clients to the model with lowest loss, requiring knowledge of $K$ and $\mathcal{O}(K)$ communication per round.
CFL~\cite{sattler2020cfl} recursively bipartitions clients based on cosine similarity of gradient updates.
FedEM~\cite{marfoq2021fedem} fits a mixture of distributions but requires maintaining multiple model components per client.
A common assumption is that heterogeneity has a discrete cluster structure, which fails when heterogeneity varies along continuous axes or has multiple simultaneous dimensions.

\textbf{Personalized federated learning.}
Ditto~\cite{li2021ditto} learns a global model and then fine-tunes per-client models with a proximal regularizer, alternating between global and local updates.
Per-FedAvg~\cite{fallah2020perfedavg} uses MAML-style~\cite{finn2017maml} meta-learning to train a global model that adapts quickly to each client.
FedProx~\cite{li2020fedprox} adds a proximal term to the local objective to prevent client drift from the global model.
Federated multi-task learning~\cite{smith2017fedmtl} explicitly models task relationships via a shared structure.
These approaches require per-client model components or adaptation steps at inference time, scaling with the number of clients.
Closer to our framing, pFedFDA~\cite{mclaughlin2024pfedfda} also conditions personalization on each client's feature distribution, but estimates a class-conditional Gaussian generative classifier rather than a fixed PCA-eigenvalue summary.
See Tan et al.~\cite{tan2022pflsurvey} for a comprehensive survey.

\textbf{Decentralized federated learning.}
Decentralized SGD~\cite{lian2017decentralized} replaces the central server with peer-to-peer communication over a graph topology, with convergence guarantees under various mixing conditions~\cite{perazzone2022communication}.
DAC~\cite{zec2022dac} extends decentralized learning with adaptive aggregation weights: each client adjusts how much it averages with each neighbor based on model similarity, enabling implicit clustering without central coordination.
Our gossip baseline uses random pairwise model averaging without such adaptation.

\textbf{Our approach} differs from all the above by not discovering client relationships. Instead, it \emph{characterizes} each client's distribution from locally-computed statistics and conditions a single shared model on it.
Most closely related is our previous work~\cite{brannvall2025conditioning}, which introduced statistics-conditioning in FL with three architectures (linear, ensemble regression, MLP) on synthetic tasks and EMNIST with three clusters.
The present work uses a simplified concatenation architecture, PCA eigenvalues rather than eigenvectors, a larger evaluation across four heterogeneity types, four datasets, and seven baselines, and the evidence that continuous statistics can outperform ground-truth cluster IDs.

\section{Method}
\label{sec:method}

\begin{figure*}
\centering
\begin{tikzpicture}[
  >=Stealth,
  modelbox/.style={draw=gray!70, thick, rounded corners=3pt, minimum height=7mm, font=\small, inner sep=4pt},
  clientbox/.style={draw=gray!70, thick, rounded corners=3pt, minimum height=7mm, minimum width=10mm, font=\footnotesize, inner sep=3pt},
  paneltitle/.style={font=\normalsize\bfseries},
  pannote/.style={font=\footnotesize\bfseries, align=center},
  commtext/.style={font=\footnotesize\bfseries, align=center},
  panelbg/.style={draw=orange!40, rounded corners=6pt, fill=orange!3, inner sep=6pt},
]
\def\pw{6.0}\def\pg{0.6}
\def\cy{0.5}\def\my{2.5}
\def\rg{4.8}
\pgfmathsetmacro{\shiftcol}{\pw+\pg}
\pgfmathsetmacro{\shiftrow}{-\rg}

\begin{scope}
  \node[panelbg, minimum width=5.8cm, minimum height=3.2cm] at (\pw/2, 1.45) {};
  \node[paneltitle] at (\pw/2, 3.4) {(a) FedAvg};
  \node[modelbox, fill=orange!15, very thick] (ma) at (\pw/2, \my) {$\boldsymbol{\theta}$};
  \node[clientbox, fill=red!12] (ca1) at (0.8, \cy) {$\boldsymbol{\theta}$};
  \node[clientbox, fill=red!12] (ca2) at (2.1, \cy) {$\boldsymbol{\theta}$};
  \node[clientbox, fill=green!12] (ca3) at (3.9, \cy) {$\boldsymbol{\theta}$};
  \node[clientbox, fill=green!12] (ca4) at (5.2, \cy) {$\boldsymbol{\theta}$};
  \draw[->, thick, orange!70] (ma) -- (ca1);
  \draw[->, thick, orange!70] (ma) -- (ca2);
  \draw[->, thick, orange!70] (ma) -- (ca3);
  \draw[->, thick, orange!70] (ma) -- (ca4);
  \node[commtext] at (\pw/2, 1.55) {global model updates};
  \node[pannote, text width=5.4cm] at (\pw/2, -0.55) {one shared model $f(\mathbf{x};\boldsymbol{\theta})$ — no personalization};
\end{scope}

\begin{scope}[shift={(\shiftcol, 0)}]
  \node[panelbg, minimum width=5.8cm, minimum height=3.2cm] at (\pw/2, 1.45) {};
  \node[paneltitle] at (\pw/2, 3.4) {(b) Clustered FL};
  \node[modelbox, fill=red!15] (mb1) at (1.4, \my) {$\boldsymbol{\theta}_1$};
  \node[modelbox, fill=green!15] (mb2) at (4.6, \my) {$\boldsymbol{\theta}_K$};
  \node[font=\small] at (\pw/2, \my) {$\cdots$};
  \node[clientbox, fill=red!12] (cb1) at (0.8, \cy) {$\boldsymbol{\theta}_1$};
  \node[clientbox, fill=red!12] (cb2) at (2.1, \cy) {$\boldsymbol{\theta}_1$};
  \node[clientbox, fill=green!12] (cb3) at (3.9, \cy) {$\boldsymbol{\theta}_K$};
  \node[clientbox, fill=green!12] (cb4) at (5.2, \cy) {$\boldsymbol{\theta}_K$};
  \draw[dashed, gray, thick] (\pw/2, \cy-0.55) -- (\pw/2, \cy+0.55);
  \draw[->, thick, red!50] (mb1) -- (cb1);
  \draw[->, thick, red!50] (mb1) -- (cb2);
  \draw[->, thick, green!50] (mb2) -- (cb3);
  \draw[->, thick, green!50] (mb2) -- (cb4);
  \node[commtext] at (\pw/2, 1.55) {$K$ models + cluster assignments};
  \node[pannote, text width=5.4cm] at (\pw/2, -0.55) {$K$ cluster models $f(\mathbf{x};\boldsymbol{\theta}_k)$ — iterative discovery};
\end{scope}

\begin{scope}[shift={(0, \shiftrow)}]
  \node[panelbg, minimum width=5.8cm, minimum height=3.2cm] at (\pw/2, 1.45) {};
  \node[paneltitle] at (\pw/2, 3.4) {(c) Personalized FL};
  \node[modelbox, fill=orange!15, very thick] (mc) at (\pw/2, \my) {$\boldsymbol{\theta}$};
  \node[clientbox, fill=red!12] (cc1) at (0.8, \cy) {$\boldsymbol{\theta}_1^{\text{p}}$};
  \node[clientbox, fill=red!12] (cc2) at (2.1, \cy) {$\boldsymbol{\theta}_2^{\text{p}}$};
  \node[clientbox, fill=green!12] (cc3) at (3.9, \cy) {$\boldsymbol{\theta}_3^{\text{p}}$};
  \node[clientbox, fill=green!12] (cc4) at (5.2, \cy) {$\boldsymbol{\theta}_N^{\text{p}}$};
  \draw[->, thick, orange!70] (mc) -- (cc1);
  \draw[->, thick, orange!70] (mc) -- (cc2);
  \draw[->, thick, orange!70] (mc) -- (cc3);
  \draw[->, thick, orange!70] (mc) -- (cc4);
  \node[commtext] at (\pw/2, 1.55) {model + per-client updates};
  \node[pannote, text width=5.4cm] at (\pw/2, -0.55) {$N$ personal models $f(\mathbf{x};\boldsymbol{\theta}_i^{\text{p}})$ — per-client fine-tuning};
\end{scope}

\begin{scope}[shift={(\shiftcol, \shiftrow)}]
  \node[panelbg, minimum width=5.8cm, minimum height=3.2cm] at (\pw/2, 1.45) {};
  \node[paneltitle] at (\pw/2, 3.4) {(d) Conditional (Ours)};
  \node[modelbox, fill=orange!15, very thick] (md) at (\pw/2, \my) {$\boldsymbol{\theta}$};
  \node[clientbox, fill=red!12] (cd1) at (0.8, \cy) {$\boldsymbol{\theta}, \mathbf{s}_1$};
  \node[clientbox, fill=red!12] (cd2) at (2.1, \cy) {$\boldsymbol{\theta}, \mathbf{s}_2$};
  \node[clientbox, fill=green!12] (cd3) at (3.9, \cy) {$\boldsymbol{\theta}, \mathbf{s}_3$};
  \node[clientbox, fill=green!12] (cd4) at (5.2, \cy) {$\boldsymbol{\theta}, \mathbf{s}_N$};
  \draw[->, thick, orange!70] (md) -- (cd1);
  \draw[->, thick, orange!70] (md) -- (cd2);
  \draw[->, thick, orange!70] (md) -- (cd3);
  \draw[->, thick, orange!70] (md) -- (cd4);
  \node[commtext] at (\pw/2, 1.55) {global model updates};
  \node[pannote, text width=5.4cm] at (\pw/2, -0.55) {one shared model $f(\mathbf{x};\boldsymbol{\theta},\mathbf{s}_i)$ — zero extra communication};
\end{scope}
\end{tikzpicture}
\caption{Four FL paradigms under data heterogeneity. (a)~FedAvg: one global model, no personalization. (b)~Clustered: $K$ separate models with iterative cluster discovery. (c)~Personalized: $N$ per-client models fine-tuned from a shared model. (d)~Ours: a single shared model conditioned on locally-computed statistics~$\mathbf{s}_i$, requiring no cluster discovery and no additional communication.}
\label{fig:paradigms}
\end{figure*}

Figure~\ref{fig:paradigms} contrasts our approach with the two dominant paradigms for handling data heterogeneity in federated learning.
Our method has three components: (1)~computing a compact distributional fingerprint for each client, (2)~conditioning a standard neural network on this fingerprint, and (3)~training the conditioned model on pooled data.
Figure~\ref{fig:pipeline} illustrates the resulting pipeline for a single client.
We describe each component in turn. Full details with all experiments are provided in the appendix.

\subsection{Client Distribution Statistics}
\label{sec:stats}

Each client $i$ computes a compact representation of its local training data distribution.
Given training set $\mathcal{D}_i^{\text{train}} = \{(\mathbf{x}, y)\}$ with $n_i = |\mathcal{D}_i^{\text{train}}|$ samples, the client constructs an augmented feature matrix by concatenating input features with one-hot encoded labels:
\begin{equation}
\mathbf{Z}_i = \bigl[\phi(\mathbf{x}) \,\|\, \text{onehot}(y);\quad \forall\, (\mathbf{x},y) \in \mathcal{D}_i^{\text{train}} \bigr]
\label{eq:concat}
\end{equation}
where $\phi(\cdot)$ is a feature extractor, $\|$ denotes concatenation, and $d = \dim(\phi(\mathbf{x})) + C$ is the combined feature-label dimension;
such that we have $\mathbf{Z}_i \in \mathbb{R}^{n_i \times d}$.
For MNIST and Fashion-MNIST, $\phi$ is the identity on flattened pixels, giving $d = 784 + C$ where $C$ is the number of classes.
For CIFAR-10 and CIFAR-100, where raw pixels are high-dimensional and less semantically meaningful, $\phi$ maps images through a cross-trained convolutional encoder to a 128-dimensional embedding, giving $d = 128 + C$.

The encoder is used only to compute each client's aggregate PCA statistics---a single 32-dimensional vector per client.
Individual sample embeddings are not used in the federated training pipeline, and at test time each client simply reuses its precomputed statistics vector.

The inclusion of one-hot labels in the feature matrix is deliberate: it ensures that the PCA eigenvalues capture not only the structure of the input features but also how they relate to the classes that are present.
This makes the resulting statistics sensitive to label shift (different class compositions), covariate shift (different input distributions), and concept shift (different label assignments), rather than just one type of variation.

The client performs PCA on $\mathbf{Z}_i$ and retains the top-$l$ eigenvalues as the statistics vector $\mathbf{s}_i \in \mathbb{R}^{l}$; we use $l{=}32$ throughout (our experiments with $l{=}8$ yielded similar results, suggesting the method is not sensitive to this choice).
This statistics vector serves as a distributional fingerprint of client $i$'s training data.
Since only the top eigenvalues are needed, iterative methods (e.g., randomized SVD) compute them in $\mathcal{O}(n_i d)$ time, scaling linearly in the number of samples and feature dimension.
This computation requires no collaboration between clients---each client independently applies an agreed-upon deterministic procedure (PCA on a shared feature representation) to its own data.

\textbf{Why eigenvalues?}
Eigenvalues capture the \emph{magnitude} of variation along each principal direction without revealing the directions.
This makes them invariant to rotations of the data matrix (up to reordering) and more robust than raw covariance statistics.
Eigenvalues from PCA on the joint feature-label matrix distinguish clients that see different classes, different feature distributions, or different label-to-feature mappings---the axes of variation that define heterogeneity in FL.
This differs from PerPCA~\cite{shi2024perpca}, a federated PCA method that learns shared and client-specific principal components for representation learning; we instead use the top eigenvalues as a fixed conditioning vector for a shared classifier.

\subsection{Conditional Model Architecture}
\label{sec:arch}

We condition a standard convolutional neural network (CNN) on the client statistics by concatenating $\mathbf{s}_i$ with the flattened feature vector after the convolutional layers and before the fully-connected (FC) classifier.
This lets the FC layers learn how to use the statistics to modulate classification.

For $28{\times}28$ grayscale inputs (MNIST, Fashion-MNIST), the architecture is:
\begin{align}
&\text{Conv}(1{\to}32, 3{\times}3) \to \text{ReLU} \to \text{MaxPool}(2) \notag \\
&\to \text{Conv}(32{\to}64, 3{\times}3) \to \text{ReLU} \to \text{MaxPool}(2) \notag \\
&\to \text{Flatten} \to [\mathbf{f}_{3136} \,\|\, \mathbf{s}_{32}] \notag \\
&\to \text{FC}(3168 {\to} 128) \to \text{ReLU} \to \text{FC}(128 {\to} C)
\end{align}

For $32{\times}32$ RGB inputs (CIFAR-10, CIFAR-100), the architecture uses four convolutional layers with batch normalization: 
\begin{align}
&\text{Conv}(3{\to}32)^2 \to \text{MaxPool}(2) \to \text{Conv}(32{\to}64)^2  \notag \\
&\to \text{MaxPool}(2) \to \text{Flatten} \to [\mathbf{f}_{4096} \,\|\, \mathbf{s}_{32}] \notag \\
&\to \text{FC}(4128 {\to} 256) \to \text{ReLU} \to \text{FC}(256 {\to} C)
\end{align}

The statistics vector adds only $32 \times H$ parameters to the first FC layer weight matrix (where $H$ is the hidden dimension), a negligible overhead: 4{,}096 parameters out of ${\sim}426{,}000$ total for the MNIST architecture ($<1\%$).

An important design choice is \emph{where} to inject the statistics.
We concatenate at the transition from convolutional to fully-connected layers because the convolutional features are shared across all clients (capturing general visual patterns), while the FC classifier adapts to each client's task through the conditioning information.
We find this simple concatenation approach works well across all experiments.

\subsection{Training and Inference}
\label{sec:training}

\textbf{Training.}
Each sample $(\mathbf{x}, y) \in \mathcal{D}_i^{\text{train}}$ is augmented with client $i$'s statistics to form a training triple $(\mathbf{x}, \mathbf{s}_i, y)$.
All clients' augmented data are concatenated and the model is trained on the pooled set.
We use centralized pooling to isolate the conditioning effect; in deployment, clients would train locally and aggregate via standard federated averaging.
The model is trained end-to-end with SGD (learning rate 0.01, momentum 0.9, batch size 64, 20~epochs).
Cross-entropy loss is used throughout.
No data augmentation is applied.

\textbf{Inference.}
At test time, for each $(\mathbf{x}, y) \in \mathcal{D}_i^{\text{test}}$, client $i$ uses its precomputed statistics vector $\mathbf{s}_i$ together with the shared model weights $\boldsymbol{\theta}$.
The client-specific prediction function is $f(\mathbf{x}; \boldsymbol{\theta}, \mathbf{s}_i)$, where $\boldsymbol{\theta}$ is shared and $\mathbf{s}_i$ is private.
No fine-tuning or adaptation steps are required at inference time.

\textbf{Privacy and communication.}
The statistics vector $\mathbf{s}_i$ is computed locally and never leaves client $i$.
The method requires the same communication as FedAvg---one model upload and one model download per round---with no cluster assignments, similarity scores, or client statistics transmitted.

\section{Experimental Setup}
\label{sec:experiments}

\begin{table*}
\centering
\caption{E1: Label shift accuracy. Conditional matches Oracle across all dataset--$K$ combinations. FedAvg collapses at high $K$ while Conditional maintains Oracle-level performance.}
\label{tab:e1_ksweep}
\setlength{\tabcolsep}{2.3pt}
\footnotesize
\begin{tabular}{@{}l cccc cccc cccc@{}}
\toprule
& \multicolumn{4}{c}{MNIST} & \multicolumn{4}{c}{Fashion-MNIST} & \multicolumn{4}{c}{CIFAR-10} \\
\cmidrule(lr){2-5} \cmidrule(lr){6-9} \cmidrule(lr){10-13}
Method & \scriptsize$K{=}2$ & \scriptsize$K{=}3$ & \scriptsize$K{=}5$ & \scriptsize$K{=}10$ & \scriptsize$K{=}2$ & \scriptsize$K{=}3$ & \scriptsize$K{=}5$ & \scriptsize$K{=}10$ & \scriptsize$K{=}2$ & \scriptsize$K{=}3$ & \scriptsize$K{=}5$ & \scriptsize$K{=}10$ \\
\midrule
Local   & .991 & .992 & .997 & 1.00 & .908 & .920 & .945 & 1.00 & .767 & .784 & .856 & 1.00 \\
FedAvg  & .986 & .978 & .933 & .522 & .902 & .886 & .745 & .412 & .735 & .554 & .325 & .172 \\
Gossip  & .982 & .983 & .892 & .560 & .895 & .899 & .825 & .560 & .745 & .755 & .533 & .566 \\
Ditto   & .993 & .994 & .996 & 1.00 & .910 & .915 & .933 & .999 & .826 & .822 & .843 & 1.00 \\
IFCA    & .995 & .995 & .999 & 1.00 & .922 & .932 & .945 & 1.00 & .847 & .766 & .906 & 1.00 \\
DAC     & .995 & .995 & .998 & 1.00 & .922 & .931 & .950 & 1.00 & .844 & .859 & .898 & 1.00 \\
\midrule
Oracle  & .996 & .996 & .999 & 1.00 & .937 & .945 & .962 & 1.00 & .861 & .886 & .927 & 1.00 \\
\best{Cond} & \best{.996} & \best{.996} & \best{1.00} & \best{1.00} & \best{.936} & \best{.942} & \best{.964} & \best{1.00} & \best{.860} & \best{.875} & \best{.929} & \best{1.00} \\
\bottomrule
\end{tabular}
\end{table*}

\subsection{Heterogeneity Taxonomy}
\label{sec:taxonomy}

We organize our evaluation around four types of data heterogeneity, each creating fundamentally different challenges.

\textbf{E1: Label shift.}
Each cluster of clients receives data from disjoint subsets of classes.
With $K$ clusters, the original $C$-class dataset is partitioned into $K$ superclass groups using a deterministic mapping (e.g., with $K{=}2$ on CIFAR-10, one cluster receives vehicles and the other animals).
We sweep $K \in \{2, 3, 5, 10\}$ for MNIST, Fashion-MNIST, and CIFAR-10, and test $K{=}20$ on CIFAR-100 using its official superclass structure.
For each $K$, we also sweep data sparsity by varying the number of clients per cluster from 5 (rich, ${\sim}6{,}000$ samples/client) to 100 (super sparse, ${\sim}200$ samples/client).
This yields 16 dataset--$K$ combinations $\times$ 5 sparsity levels for the main datasets.

\textbf{E2: Covariate shift.}
All clusters predict the same superclass labels but observe different input distributions.
In E2a (subclass covariate shift), clusters see different subclasses: e.g., with $K{=}2$ on CIFAR-10, both clusters predict vehicle vs.\ animal, but one sees trucks and dogs while the other sees cars and cats.
In E2b (rotation covariate shift), different clusters see images rotated by different angles (e.g., $0\degree$ vs.\ $180\degree$).
Each client's test set matches its training distribution (local evaluation).

\textbf{E3: Concept shift.}
Different clusters apply different labeling functions to the same images.
In E3a (semantic concept shift), clusters use different semantic groupings---e.g., one MNIST cluster uses odd/even labels while another uses below-5/above-5.
In E3b (label permutation), all clusters perform 10-class classification but each uses a different random permutation of the label mapping.
FedAvg is particularly destructive here because averaging models with incompatible label semantics produces a model that satisfies none of them.

\textbf{E4: Domain shift and combined heterogeneity.}
E4a tests cross-dataset domain shift (MNIST clients coexist with Fashion-MNIST clients, both performing 10-class classification on their respective domains).
E4b stacks concept shift with covariate shift on CIFAR-10: two concept groups (semantic vs.\ size classification) are each subdivided into $C \in \{2,3,4\}$ covariate clusters with different subclass distributions, yielding $2C$ total clusters.
This is the most challenging setting because heterogeneity is inherently multi-dimensional.

\subsection{Datasets and Data Splits}

We use four image classification datasets:
\textbf{MNIST}~\cite{lecun2010mnist} (60K images, 28$\times$28, 10 digit classes),
\textbf{Fashion-MNIST}~\cite{xiao2017fmnist} (60K images, 28$\times$28, 10 clothing classes),
\textbf{CIFAR-10}~\cite{krizhevsky2009cifar} (50K images, 32$\times$32$\times$3, 10 object classes), and
\textbf{CIFAR-100}~\cite{krizhevsky2009cifar} (50K images, 32$\times$32$\times$3, 100 fine-grained classes in 20 superclasses). All counts are training set sizes; each dataset also provides 10K test images.

For each experiment configuration, clients within a cluster share the same data distribution but receive disjoint subsets of the cluster's data.
The number of clients per cluster determines data sparsity.
Test data is split identically to training data: each client is evaluated on test samples matching its own distribution (local test sets).

\subsection{Baselines}

We compare against seven baselines spanning all major FL paradigms:
\begin{itemize}
\item \textbf{Local}: Each client trains independently on its own data. This is a lower bound showing what is achievable without collaboration.
\item \textbf{FedAvg}~\cite{mcmahan2017fedavg}: Standard federated averaging. A single global model is trained by averaging local updates across all clients.
\item \textbf{Gossip}: Decentralized learning with random pairwise model averaging on a random topology, without adaptive aggregation weights.
\item \textbf{Oracle}: Trains one separate model per ground-truth cluster, pooling data within each cluster. This is the ceiling for clustering-based methods since it uses perfect cluster assignments.
\item \textbf{IFCA}~\cite{ghosh2020ifca}: Iterative Federated Clustering Algorithm. Maintains $K$ model hypotheses; each client selects the best-fitting model based on training loss.
\item \textbf{DAC}~\cite{zec2022dac}: Decentralized Adaptive Clustering. Clients average models with neighbors, weighting by cosine similarity of model parameters.
\item \textbf{Ditto}~\cite{li2021ditto}: Each client maintains a local model that alternates updates between global averaging and local fine-tuning with proximal regularization ($\lambda{=}1.0$).
\end{itemize}

For IFCA and DAC, we provide the true number of clusters $K$; an advantage that would be unavailable in practice.

\subsection{Training Protocol}

All methods use the same CNN backbone architecture described in Sec.~\ref{sec:arch}, without the statistics concatenation for non-conditional methods.
Training uses SGD with learning rate 0.01, momentum 0.9, batch size 64, and 20~epochs.
For Ditto, we use the alternating-updates implementation from Li et al.~\cite{li2021ditto} with proximal parameter $\lambda{=}1.0$.
For IFCA, we use 5~rounds of cluster refinement.
All experiments use 32 PCA components for client statistics.
Each configuration uses deterministic data splits; variance across random seeds is small ($< 0.5\%$) so we report single-run results.

\section{Results}
\label{sec:results}

We organize results around three empirical claims, each supported by multiple experiments.
Across all 97~configurations tested, our Conditional method outperforms or matches the Oracle in 95 (98\%).
The two exceptions are CIFAR-100 $K{=}20$ (3.5\% below Oracle) and CIFAR-10 E3a at super-sparse (8.5\% below Oracle).
The CIFAR-100 gap reflects the fine-grained difficulty of distinguishing 20 superclasses---Conditional still achieves the best non-Oracle result (65.2\%, vs.\ 65.1\% for DAC). The E3a gap occurs only at super-sparse (${\sim}200$ samples/client) on CIFAR-10, where eigenvalue estimates from learned embeddings become noisier.

\subsection{Claim 1: Matches Oracle Across Heterogeneity Types}
\label{sec:claim1}

Table~\ref{tab:summary} shows representative results from each of the seven experiment types.
In every case, Conditional achieves the highest accuracy among non-Oracle methods and matches Oracle within 1\%.
This is notable because Conditional has no access to cluster assignments---it operates from distributional statistics alone.

Table~\ref{tab:e1_ksweep} shows the full E1 label shift results across three datasets and $K \in \{2,3,5,10\}$.
Several patterns emerge.
First, \textbf{FedAvg degrades catastrophically} with increasing $K$: on CIFAR-10, accuracy drops from 73.5\% ($K{=}2$) to 17.2\% ($K{=}10$), because global averaging cannot reconcile 10 incompatible local objectives.
Second, \textbf{Conditional tracks Oracle} at every operating point, with gaps below 1.1\% everywhere.
Third, IFCA and DAC are competitive baselines on average but show occasional instability---IFCA drops to 76.6\% on CIFAR-10 $K{=}3$ (vs.\ 88.6\% for Oracle), likely due to cluster estimation failure.
Fourth, all methods converge to 100\% at $K{=}10$ on the three 10-class datasets, since each cluster then contains a single class (trivial task).

The pattern extends to other heterogeneity types (full tables in the appendix).
Under covariate shift (E2a), where all clusters predict the same labels but see different subclasses, FedAvg remains reasonable (since the task is shared) but Conditional still matches Oracle exactly (e.g., both at 97.6\% on CIFAR-10 $K{=}2$).
Under rotation-based covariate shift (E2b), Conditional matches Oracle within 0.5\% on MNIST and \emph{exceeds} it on two FMNIST configurations (by 0.1\% and 0.4\%).
Under semantic concept shift (E3a), Conditional matches Oracle on MNIST and FMNIST at the richest data setting (99.4\% vs.\ 99.3\% on MNIST, 95.0\% vs.\ 95.0\% on FMNIST), while FedAvg drops to 79.3\% and 67.1\% respectively.
Under cross-dataset domain shift (E4a), where MNIST and Fashion-MNIST clients coexist, Conditional again matches Oracle (95.5\% vs.\ 95.7\%).

\subsection{Claim 2: Beats Oracle on Combined Heterogeneity}
\label{sec:claim2}

When heterogeneity has multiple simultaneous dimensions, discrete cluster assignments become fundamentally insufficient.
Consider E4b: a client is characterized by \emph{both} its concept group (semantic vs.\ size classification) and its covariate cluster (which subclasses it sees).
The Oracle baseline assigns one cluster ID per client, which is a single integer encoding both dimensions.
In contrast, the 32-dimensional continuous statistics vector can separately encode information about the concept axis and the covariate axis, providing a richer representation.

\begin{table}
\centering
\caption{E4b: Combined heterogeneity on CIFAR-10. Conditional consistently \emph{exceeds} Oracle because continuous statistics capture multi-dimensional variation that a discrete cluster ID cannot. $C$ = covariate clusters per concept group; total clusters = $2C$.}
\label{tab:e4b}
\setlength{\tabcolsep}{3.5pt}
\footnotesize
\begin{tabular}{@{}lcccc@{}}
\toprule
Method & $C{=}2$ (4 cl.) & $C{=}3$ (6 cl.) & $C{=}4$ (8 cl.) & Mean \\
\midrule
Local      & .818 & .854 & .886 & .853 \\
FedAvg     & .754 & .743 & .764 & .754 \\
Gossip     & .775 & .789 & .786 & .783 \\
Ditto      & .868 & .880 & .906 & .885 \\
IFCA       & .752 & .740 & .892 & .795 \\
DAC        & .888 & .854 & .896 & .879 \\
\midrule
Oracle     & .891 & .914 & .939 & .915 \\
\best{Cond} & \best{.920} & \best{.935} & \best{.955} & \best{.937} \\
\midrule
\multicolumn{1}{@{}r}{\scriptsize$\Delta$(Cond$-$Oracle)} & \scriptsize+2.9\% & \scriptsize+2.2\% & \scriptsize+1.5\% & \scriptsize+2.2\% \\
\bottomrule
\end{tabular}
\end{table}

Table~\ref{tab:e4b} shows E4b results on CIFAR-10 as we increase the number of covariate clusters per concept group ($C$) from 2 to 4.
Conditional \textbf{exceeds Oracle by 1.5--2.9 percentage points} across all configurations, with a mean advantage of 2.2\%.
The gap ranges from 2.9\% at $C{=}2$ (4 total clusters) to 1.5\% at $C{=}4$ (8 clusters).

\begin{figure*} 
\centering
\includegraphics{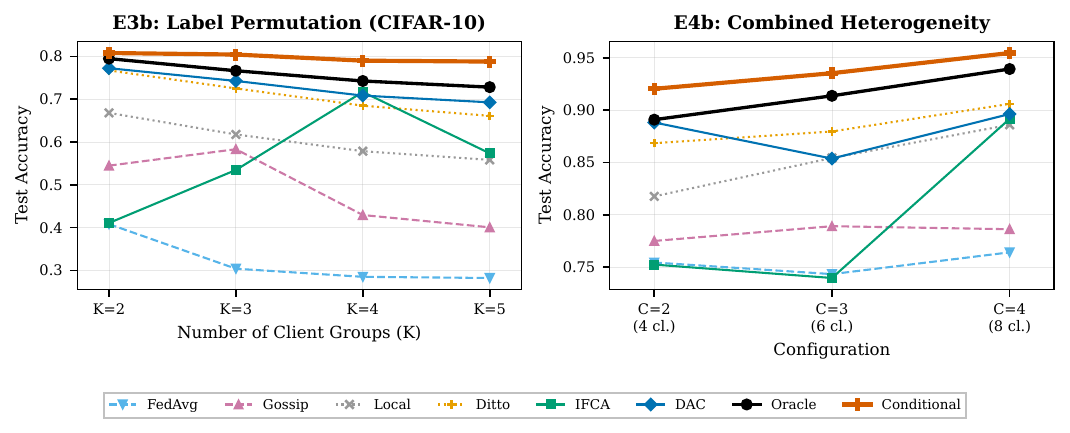}
\caption{Conditional exceeds Oracle on complex heterogeneity. Left: E3b label permutation on CIFAR-10 ($K$ sweep). Right: E4b combined heterogeneity (concept + covariate shift). Continuous statistics provide richer conditioning than discrete cluster IDs.}
\label{fig:beats_oracle}
\end{figure*}

Figure~\ref{fig:beats_oracle} visualizes this pattern across two experiment types.
In the left panel (E3b label permutation on CIFAR-10), Conditional exceeds Oracle at every value of $K$, with the largest gap (+1.3\%) at $K{=}2$.
Note that IFCA shows extreme instability on this experiment: it achieves only 41.1\% at $K{=}2$ (well below other methods) but jumps to 71.7\% at $K{=}4$, likely due to fortuitous cluster initialization.
DAC, while more stable, falls 2--4\% below Oracle.
In the right panel (E4b combined heterogeneity), Conditional is the clear top performer, exceeding Oracle at every configuration.

IFCA fails on E4b at $C{=}2$ and $C{=}3$ (75.2\% and 74.0\%, at or below FedAvg) but recovers at $C{=}4$ (89.2\%).
This pattern is consistent with IFCA's sensitivity to initialization: its loss-based E-step conflates concept and covariate differences into a single scalar, and with fewer, more overlapping clusters ($C{=}2{,}3$), random initialization is unlikely to land near a good assignment.
At $C{=}4$ each cluster is more homogeneous, making the loss landscape more separable.
DAC provides the strongest baseline on E4b (mean 87.9\%), likely because its soft similarity-based aggregation handles multi-dimensional heterogeneity better than IFCA's hard assignments, but it still falls 5.8\% below Conditional.

\begin{table*}
\centering
\caption{E3b: Label permutation ($K$ sweep). Conditional exceeds Oracle on all three datasets, with the largest gains on CIFAR-10.}
\label{tab:e3b}
\setlength{\tabcolsep}{2.3pt}
\footnotesize
\begin{tabular}{@{}l cccc cccc cccc@{}}
\toprule
& \multicolumn{4}{c}{MNIST} & \multicolumn{4}{c}{Fashion-MNIST} & \multicolumn{4}{c}{CIFAR-10} \\
\cmidrule(lr){2-5} \cmidrule(lr){6-9} \cmidrule(lr){10-13}
Method & \scriptsize$K{=}2$ & \scriptsize$K{=}3$ & \scriptsize$K{=}4$ & \scriptsize$K{=}5$ & \scriptsize$K{=}2$ & \scriptsize$K{=}3$ & \scriptsize$K{=}4$ & \scriptsize$K{=}5$ & \scriptsize$K{=}2$ & \scriptsize$K{=}3$ & \scriptsize$K{=}4$ & \scriptsize$K{=}5$ \\
\midrule
Local   & .981 & .977 & .973 & .969 & .877 & .863 & .854 & .845 & .668 & .618 & .579 & .558 \\
FedAvg  & .495 & .332 & .295 & .306 & .454 & .330 & .278 & .257 & .408 & .304 & .285 & .282 \\
Gossip  & .691 & .798 & .568 & .585 & .619 & .677 & .477 & .484 & .545 & .583 & .429 & .401 \\
Ditto   & .982 & .980 & .975 & .970 & .877 & .854 & .833 & .815 & .767 & .725 & .685 & .661 \\
IFCA    & .990 & .985 & .982 & .844 & .892 & .593 & .659 & .705 & .411 & .535 & .717 & .574 \\
DAC     & .989 & .986 & .982 & .979 & .891 & .878 & .866 & .856 & .773 & .743 & .709 & .693 \\
\midrule
Oracle  & .992 & .988 & .988 & .986 & .914 & .904 & .899 & .894 & .795 & .767 & .743 & .728 \\
\best{Cond} & \best{.992} & \best{.991} & \best{.991} & \best{.989} & \best{.916} & \best{.914} & \best{.911} & \best{.910} & \best{.808} & \best{.804} & \best{.790} & \best{.788} \\
\bottomrule
\end{tabular}
\end{table*}

Table~\ref{tab:e3b} supports this claim with E3b label permutation results.
Here, Conditional exceeds Oracle on \emph{all three datasets}:
by about 0.3\% on MNIST,
by 0.2--1.6\% on Fashion-MNIST,
and by 1.3--6.0\% on CIFAR-10 (where the advantage is largest).
The CIFAR-10 results are striking: at $K{=}5$, Conditional achieves 78.8\% while Oracle reaches only 72.8\%---a full 6\% gap.
This confirms that when label semantics differ across clients, continuous statistics provide the model with richer conditioning than knowing which of $K$ permutations was applied.

\subsection{Claim 3: Uniquely Sparsity-Robust}
\label{sec:claim3}

Data sparsity is a critical concern in FL: real-world clients often have limited local data.
Clustering methods require sufficient data to estimate cluster membership; personalization methods need enough to fit per-client parameters.
Both degrade with only a few hundred samples per client.

\begin{figure*} 
\centering
\includegraphics
{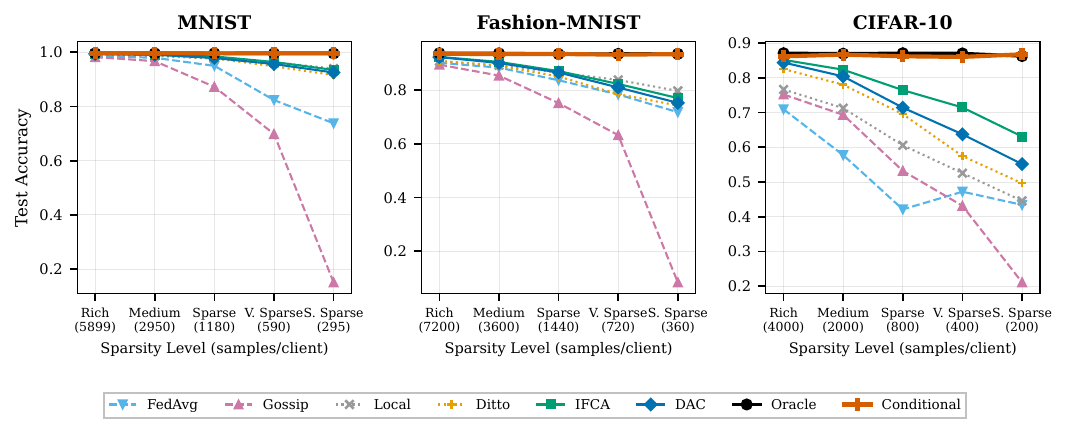}
\caption{Sparsity robustness on E1 Label Shift ($K{=}2$). Conditional and Oracle maintain flat accuracy as data decreases 20-fold (from ${\sim}6{,}000$ to ${\sim}200$ samples/client). All other methods degrade, with Gossip collapsing to near-random.}
\label{fig:sparsity}
\end{figure*}

Figure~\ref{fig:sparsity} shows E1 label shift performance as client data decreases from ``Rich'' (${\sim}4{,}000$--$6{,}000$ samples) to ``Super Sparse'' (${\sim}200$--$300$ samples).
Two lines are essentially flat: Oracle and Conditional.
All other methods show clear downward trends, with Gossip collapsing catastrophically (to 15\% on MNIST, 8\% on FMNIST, 21\% on CIFAR-10).

\begin{table}
\centering
\caption{E1 sparsity sweep on CIFAR-10 ($K{=}2$). Conditional maintains near-constant accuracy across a 20$\times$ reduction in data per client. $\Delta$: relative degradation from Rich to Super Sparse.}
\label{tab:sparsity}
\setlength{\tabcolsep}{2.3pt}
\footnotesize
\begin{tabular}{@{}lcccccc@{}}
\toprule
& Rich & Med. & Sparse & V.\,Sparse & S.\,Sparse & $\Delta$ \\
& \scriptsize(4000) & \scriptsize(2000) & \scriptsize(800) & \scriptsize(400) & \scriptsize(200) & \\
\midrule
Local   & .767 & .713 & .606 & .526 & .446 & $-$41.9\% \\
FedAvg  & .709 & .577 & .421 & .472 & .434 & $-$38.8\% \\
Gossip  & .752 & .694 & .532 & .431 & .211 & $-$72.0\% \\
Ditto   & .825 & .780 & .695 & .575 & .496 & $-$39.9\% \\
IFCA    & .852 & .824 & .765 & .715 & .631 & $-$25.9\% \\
DAC     & .844 & .804 & .714 & .637 & .552 & $-$34.6\% \\
\midrule
Oracle  & .871 & .870 & .871 & .870 & .863 & $-$0.9\% \\
\best{Cond} & \best{.862} & \best{.867} & \best{.862} & \best{.860} & \best{.868} & \best{+0.7\%} \\
\bottomrule
\end{tabular}
\end{table}

Table~\ref{tab:sparsity} quantifies the degradation on CIFAR-10, the most challenging dataset.
Conditional's accuracy \emph{increases} slightly from 86.2\% (Rich) to 86.8\% (Super Sparse), a change within noise.
In contrast, IFCA loses 25.9\% of its performance, DAC loses 34.6\%, Ditto loses 39.9\%, and Gossip loses 72.0\%.
Even FedAvg, which requires no clustering, loses 38.8\%.

This sparsity robustness holds across all experiment types and datasets, though the effect is strongest under label and covariate shift.
On MNIST, Conditional maintains 99.6\% from Rich to Super Sparse (Oracle: 99.6\%).
On Fashion-MNIST, it maintains 93.5\% (Oracle: 93.6\%).
On CIFAR-10 under concept shift (E3a), it maintains 90.4\% Rich to 81.4\% Super Sparse---a modest decline, but far less than IFCA (75.9\% to 72.6\%), DAC (84.7\% to 73.5\%), or Ditto (87.8\% to 71.8\%).

The explanation is architectural: Conditional's personalization mechanism is the precomputed statistics vector, not learned client-specific parameters.
Since the statistics are fixed before training begins, the model never needs to ``discover'' client structure from scarce data.
The shared model is trained on \emph{pooled} data from all clients, so the total training set size remains constant regardless of individual client sparsity.
In effect, sparsity affects the baselines' ability to estimate client relationships, but does not affect our model's ability to condition on already-known statistics.

\section{Discussion}
\label{sec:discussion}

\textbf{Why continuous statistics outperform discrete clusters.}
Two factors explain why Conditional exceeds Oracle on complex heterogeneity (E3b: up to +6.0\%; E4b: +1.5--2.9\%) but merely matches it on simpler single-axis heterogeneity (E1).
First, the 32-dimensional statistics vector encodes the full spectral signature of each client's distribution, capturing variation along multiple axes simultaneously---information that a single cluster ID discards.
Second, Oracle trains \emph{separate} models per cluster that share no parameters, whereas Conditional trains a \emph{single} shared model that leverages commonalities across all clients while using the statistics to differentiate, gaining both the benefit of full data pooling and client-specific adaptation.
Because clients with similar data distributions produce similar statistics~$\mathbf{s}_i$, the shared model implicitly learns to treat them alike---achieving the effect of clustering without explicit cluster assignments.
Beyond an existence result for the linear case~\cite{brannvall2025conditioning}, formal analysis of the deep PCA-conditioned model is open.

\textbf{Why the method is sparsity-robust.}
Baselines degrade under sparsity because they estimate client relationships (cluster assignments, similarity weights, or per-client parameters) from scarce data.
Our method sidesteps this for two reasons.
First, PCA eigenvalues are precomputed from raw data and remain informative with a few hundred samples; top sample eigenvalues converge faster than eigenvectors, which is why we use them as client fingerprints.
Second, the shared model trains on \emph{pooled} data, so the total training set size is independent of individual client sparsity.

\textbf{When does Conditional work best?}
The largest advantages over baselines appear when (1)~heterogeneity is severe enough that FedAvg fails, (2)~the cluster structure is complex enough that IFCA's hard assignment fails, and (3)~data is sparse enough that per-client methods overfit.
IFCA's performance is particularly variable: it achieves perfect clustering (ARI${=}1.0$) on E4a domain shift but completely fails (ARI${=}0.0$) on E4b combined heterogeneity, and one cannot know \emph{a priori} whether the heterogeneity structure will be amenable to iterative clustering.
In the benign case of mild covariate shift with ample data (E2a), all methods perform similarly.
In our experiments, conditioning did not reduce performance in any configuration: the FC layers learn to ignore uninformative statistics while providing gains where other methods struggle.

\textbf{Comparison across heterogeneity types.}
Method rankings are broadly consistent across experiments: Conditional $\approx$ Oracle $>$ DAC $>$ Ditto $>$ Local $>$ FedAvg $>$ Gossip, with IFCA as the main outlier due to its sensitivity to initialization.
However, gaps vary dramatically by heterogeneity type.
Under label shift (E1), the spread is enormous: 17.2\% (FedAvg) to 100\% (Oracle/Conditional) at $K{=}10$.
Under covariate shift (E2a), all methods exceed 92\%.
Under concept shift (E3b), FedAvg collapses to 26--50\% accuracy.
This variability underscores the need for methods robust across heterogeneity types---a need that Conditional uniquely satisfies.

\textbf{Communication, computation, and privacy.}
Our method communicates the same information as FedAvg---model parameter updates and nothing else---adding no overhead beyond the $<$1\% increase in model parameters.
In contrast, Ditto doubles local computation and storage (two models per client), IFCA increases bandwidth by a factor of $K$, and DAC requires peer-to-peer model exchanges.
Because no client statistics, cluster assignments, or similarity scores are transmitted, the method is fully compatible with secure aggregation protocols and with differential privacy via DP-SGD, since standard gradient clipping and noise injection apply unchanged to the shared parameters.
Methods that discover client relationships may need specialized privacy protocols---e.g., Listo Zec et al.~\cite{listozec2024ppdl} propose a protocol combining secure aggregation with adversarial bandit optimization to protect collaboration preferences---a complexity our approach avoids.

\textbf{Limitations.}
Our evaluation uses a centralized pooling protocol for training, which simplifies the evaluation but does not capture the communication dynamics of iterative federated rounds.
A fully federated implementation would require each client to train locally with statistics as fixed conditioning, then aggregate; the statistics computation itself is purely local and adds no communication.
We leave the empirical evaluation of convergence under federated rounds to future work.
For high-dimensional inputs (CIFAR-10/100), computing PCA statistics requires a shared embedding protocol---in our experiments we use a cross-trained encoder, but in practice the federation could agree on any fixed feature extractor (e.g., a pretrained ResNet); comparing these choices empirically is future work.
For low-dimensional inputs (MNIST, Fashion-MNIST), statistics are computed directly on raw pixels with no embedding step.
Additionally, while PCA eigenvalues are effective for the types of heterogeneity studied here, they may be less informative for shifts that do not affect second-order statistics (e.g., adversarial perturbations or changes in rare class examples).

\textbf{Conditioning architecture.}
We inject client statistics by concatenating~$\mathbf{s}_i$ with the flattened CNN output before the fully-connected layers.
An alternative is to condition the convolutional layers themselves---for example using Feature-wise Linear Modulation (FiLM)~\cite{perez2018film}, which applies learned affine transformations to intermediate feature maps based on conditioning information.
We experimented with FiLM conditioning but found that it did not improve over simple concatenation in our setting; performance was slightly worse, likely because the convolutional layers benefit from being fully shared across clients while the FC layers provide sufficient capacity for client-specific adaptation.
More expressive conditioning mechanisms may prove beneficial for harder tasks or larger models, but we leave this to future work.

\textbf{Model capacity.}
Concatenating~$\mathbf{s}_i$ increases the parameter count of the first FC layer by $32 \times H$ weights, a less than 1\% increase in total model parameters (Sec.~\ref{sec:arch}).
In our previous work~\cite{brannvall2025conditioning}, we controlled for this by providing the same total parameter budget to all compared methods.
Given the modest increase, we did not equalize capacity here.
A controlled capacity comparison across all baselines remains a natural extension.

\textbf{Temporal concept shift.}
Our method computes the statistics vector once from each client's training data and treats it as fixed throughout training and inference.
This design does not account for temporal concept shift---changes in the data distribution over time---which is challenging because stale statistics may no longer reflect the client's current distribution, and because gradual drift makes it unclear when or how frequently to recompute.
For sudden regime shifts (e.g., a client's task changes abruptly), the conditioning approach offers a natural remedy: the client recomputes its statistics vector from recent data, and the shared model adapts through the existing conditioning mechanism without retraining.
However, handling gradual concept drift---where the distribution evolves slowly and the boundary between regimes is unclear---remains an open problem that we leave to future work.

\textbf{Statistics choice.}
The inclusion of one-hot labels in the PCA input matrix (Eq.~\ref{eq:concat}) is a deliberate design choice that makes the statistics sensitive to label shift, covariate shift, and concept shift simultaneously.
Removing labels would likely reduce sensitivity to label and concept shift while preserving covariate shift detection.
At the other extreme, using simpler statistics such as per-feature variance would discard cross-feature correlations, while using eigenvectors rather than eigenvalues would produce higher-dimensional and potentially less robust fingerprints.
Exploring richer statistics (higher-order moments, neural network embeddings) and systematic ablation of these design choices are natural directions for future work.

\textbf{Broader applicability.}
While we frame the method in a federated learning context, the core idea---conditioning a shared model on distributional statistics of data subsets---applies more broadly.
Any dataset collected across heterogeneous contexts (e.g., medical imaging from different hospitals or sensors) can be partitioned by metadata, with per-partition PCA statistics appended to the features. The same concatenation mechanism then lets a single model adapt across contexts without separate models or manual harmonization.

\section{Conclusion}
\label{sec:conclusion}

We have shown that conditioning a single federated model on locally-computed PCA eigenvalue statistics provides a simple and effective approach to personalized federated learning.
The key insight is that \emph{characterizing} each client's data distribution is more effective than \emph{discovering} client relationships, particularly under data sparsity or multi-dimensional heterogeneity.

Across 97~configurations spanning four types of data heterogeneity, four datasets, and five sparsity levels, our method: (1)~matches the Oracle baseline that knows true cluster assignments across all heterogeneity types; (2)~surpasses Oracle by 1--6\% on combined heterogeneity, where continuous statistics capture variation that discrete cluster IDs cannot; and (3)~is uniquely robust to data sparsity as client data decreases 20-fold, while all seven baselines degrade.

The method requires zero additional communication beyond standard federated averaging, adds less than 1\% model parameters, and needs no cluster estimation or per-client fine-tuning.
These properties make it particularly suited to practical FL deployments where heterogeneity type is unknown, data is scarce, and communication budgets are limited.
More broadly, the same conditioning approach applies to any dataset where metadata identifies distinct collection contexts.

\bibliographystyle{IEEEtran}
\bibliography{main}

\onecolumn
\appendices
\section{Preliminary Results from MCDC 2025 Workshop Paper}
\label{sec:mcdc}

The local characteristic statistics conditioning approach was first presented as a non-archival paper at the ICLR 2025 Workshop on Modular, Collaborative and Decentralized Deep Learning (MCDC)~\cite{brannvall2025conditioning}, with experiments on synthetic regression/classification tasks and EMNIST character recognition.
The FLICS 2026 paper extends this with image classification datasets (MNIST, Fashion-MNIST, CIFAR-10, CIFAR-100), seven baselines (vs.\ three reference models), a systematic heterogeneity taxonomy (label/covariate/concept/combined shift), PCA eigenvalue statistics on embeddings instead of eigenvector of raw images only, and sparsity analysis.
The original MCDC results are reproduced in this Appendix for completeness.

\subsection{Synthetic Tasks}

The MCDC paper \cite{brannvall2025conditioning} evaluated conditioning on two synthetic tasks: linear regression and logistic regression.
In both tasks, clusters of 100 clients each were generated with distinct data distributions, and the first PCA eigenvector of each client's covariance matrix served as the conditioning statistic.
Three conditioning architectures were compared:
\begin{itemize}
    \item \textbf{Conditional linear} ($\hat{y} = \mathbf{x}^T W \boldsymbol{\mu}_i$): input features are element-wise multiplied by a learned linear projection of the conditioning vector
    \item \textbf{Ensemble regression}: separate models are blended using softmax weights predicted from the conditioning vector
    \item \textbf{MLP}: conditioning vector is concatenated with input features and fed through a multi-layer perceptron
\end{itemize}

These were compared against three reference models: \emph{global} (single model for all clients), \emph{cluster} (one model per ground-truth cluster), and \emph{client} (one model per client).

\begin{table}[htbp]
\centering
\caption[MCDC synthetic tasks (3 clusters)]{MCDC synthetic tasks: 3 clusters, 100 clients each, 10 features. Lower RMSE is better for linear regression; higher accuracy is better for logistic regression. Best reference model and best conditional model in bold.}
\label{tab:mcdc_synth3}
\small
\begin{tabular}{l|ccc|ccc}
\toprule
& \multicolumn{3}{c|}{Reference models} & \multicolumn{3}{c}{Conditional models} \\
Task & Global & Cluster & Client & Ens & MLP & Lin \\
\midrule
Linreg (RMSE $\downarrow$) & 14.901 & \textbf{0.100} & 0.106 & \textbf{0.104} & 0.134 & 0.127 \\
Logreg (acc $\uparrow$)     & 0.700  & \textbf{0.997} & 0.944 & \textbf{0.989} & 0.985 & 0.964 \\
\bottomrule
\end{tabular}
\end{table}

\begin{table}[htbp]
\centering
\caption[MCDC synthetic tasks (8 clusters)]{MCDC synthetic tasks: 8 clusters, 100 clients each, 10 features.}
\label{tab:mcdc_synth8}
\small
\begin{tabular}{l|ccc|ccc}
\toprule
& \multicolumn{3}{c|}{Reference models} & \multicolumn{3}{c}{Conditional models} \\
Task & Global & Cluster & Client & Ens & MLP & Lin \\
\midrule
Linreg (RMSE $\downarrow$) & 17.655 & \textbf{0.100} & 0.107 & \textbf{0.109} & 0.162 & 0.257 \\
Logreg (acc $\uparrow$)     & 0.621  & \textbf{0.997} & 0.943 & \textbf{0.989} & 0.966 & 0.939 \\
\bottomrule
\end{tabular}
\end{table}

\textbf{Key observations:} The ensemble conditioning model performs best among the three architectures, nearly matching the cluster oracle on both tasks.
The global model degrades with more clusters (RMSE increases from 14.9 to 17.7; accuracy drops from 0.700 to 0.621), while all conditional models remain close to the cluster-level performance.

\subsection{EMNIST Character Recognition}

The MCDC paper also evaluated on EMNIST Balanced with 3 clusters: digits (0--9), uppercase letters (A--Z), and lowercase letters (a--z), with approximately 2{,}500 samples per client.
A CNN classifier was conditioned on the first PCA eigenvector of each client's data.
The conditioning vector was concatenated with the CNN's flattened feature map before the fully-connected layers (i.e., the MLP-style architecture from the synthetic experiments, adapted to a CNN backbone).

\begin{table}[htbp]
\centering
\caption[MCDC EMNIST: accuracy vs.\ number of PCA components]{MCDC EMNIST: effect of number of PCA components ($n_c$) used for conditioning. A single component suffices; additional components provide no further benefit.}
\label{tab:mcdc_components}
\begin{tabular}{c|ccc|c}
\toprule
$n_c$ & Global & Cluster & Client & Cond \\
\midrule
0 & 0.844 & 0.969 & 0.884 & --- \\
1 & 0.847 & 0.970 & 0.880 & 0.967 \\
2 & 0.848 & 0.971 & 0.878 & 0.967 \\
3 & 0.847 & 0.971 & 0.876 & 0.966 \\
4 & 0.848 & 0.971 & 0.874 & 0.966 \\
8 & 0.850 & 0.972 & 0.871 & 0.966 \\
\bottomrule
\end{tabular}
\end{table}

\textbf{Key observation:} A single PCA eigenvector is sufficient for conditioning---adding additional eigenvectors provided no measurable benefit in our experiments (Table~\ref{tab:mcdc_components}).
The conditional model (0.967) nearly matches the cluster oracle (0.970) while substantially outperforming both the global model (0.847) and per-client models (0.880).

\begin{figure}[htbp]
\centering
\includegraphics[width=\textwidth]{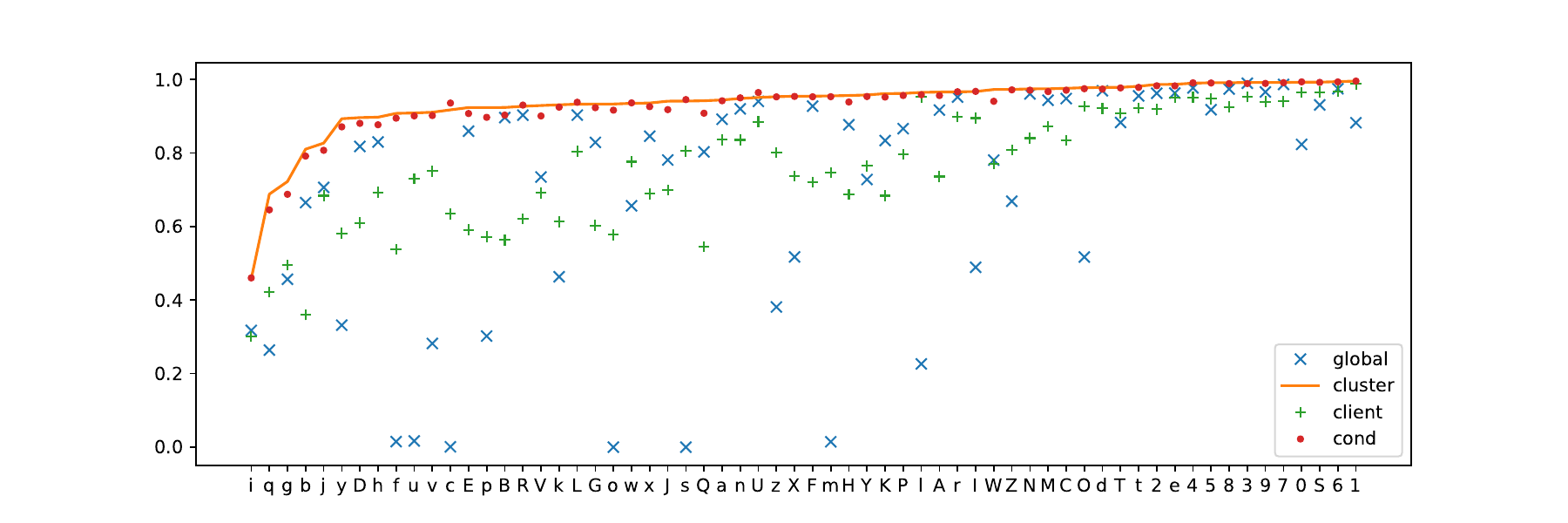}
\caption[MCDC EMNIST: per-character accuracy]{MCDC EMNIST: per-character accuracy for digits, uppercase, and lowercase character subsets (1 PCA component). The global model struggles with similar looking characters that appear in multiple subsets (e.g., lowercase `o' vs uppercase `O' vs digit `0'), while the conditional model resolves these ambiguities internally without having to identify peer clusters.}
\label{fig:mcdc_characters}
\end{figure}

Table~\ref{tab:mcdc_confusable} shows accuracy on confusable character pairs---characters whose visual appearance is shared across the digit, uppercase, and lowercase subsets.
The global model collapses to near-zero on several lowercase characters (e.g., `o': 0.0\%, `s': 0.0\%, `c': 0.1\%) because these are visually indistinguishable from their uppercase or digit counterparts without being able to infer which character subset the client uses.
The conditional model resolves this ambiguity by conditioning on client statistics, achieving accuracies close to the cluster oracle.

\begin{table}[htbp]
\centering
\caption[MCDC EMNIST: confusable character accuracy]{MCDC EMNIST: accuracy on visually confusable characters across subsets. Characters are grouped by visual similarity; the global model fails on lowercase characters with cross-subset overlap, while the conditional model resolves these using client characteristic statistics only.}
\label{tab:mcdc_confusable}
\small
\setlength{\tabcolsep}{4pt}
\begin{tabular}{cl|cccc}
\toprule
Group & Char & Global & Client & Cluster & Cond \\
\midrule
\multirow{3}{*}{o/O/0}
  & o (lower) & .000 & .578 & .933 & .917 \\
  & O (upper) & .517 & .927 & .978 & .975 \\
  & 0 (digit) & .824 & .965 & .993 & .994 \\
\midrule
\multirow{3}{*}{l/I/1}
  & l (lower) & .227 & .952 & .965 & .959 \\
  & I (upper) & .490 & .895 & .967 & .968 \\
  & 1 (digit) & .882 & .988 & .996 & .996 \\
\midrule
\multirow{3}{*}{z/Z/2}
  & z (lower) & .381 & .801 & .953 & .953 \\
  & Z (upper) & .669 & .809 & .973 & .972 \\
  & 2 (digit) & .963 & .919 & .987 & .983 \\
\midrule
\multirow{3}{*}{s/S/5}
  & s (lower) & .000 & .805 & .941 & .946 \\
  & S (upper) & .931 & .966 & .993 & .992 \\
  & 5 (digit) & .918 & .948 & .991 & .991 \\
\midrule
\multirow{2}{*}{k/K}
  & k (lower) & .464 & .613 & .931 & .925 \\
  & K (upper) & .834 & .684 & .962 & .952 \\
\midrule
\multirow{2}{*}{f/F}
  & f (lower) & .015 & .538 & .908 & .895 \\
  & F (upper) & .928 & .721 & .955 & .953 \\
\midrule
\multirow{2}{*}{p/P}
  & p (lower) & .302 & .572 & .924 & .897 \\
  & P (upper) & .866 & .797 & .962 & .957 \\
\midrule
\multirow{2}{*}{c/C}
  & c (lower) & .001 & .635 & .918 & .936 \\
  & C (upper) & .948 & .835 & .976 & .971 \\
\bottomrule
\end{tabular}
\end{table}

\begin{figure}[htbp]
\centering
\includegraphics[width=0.85\textwidth]{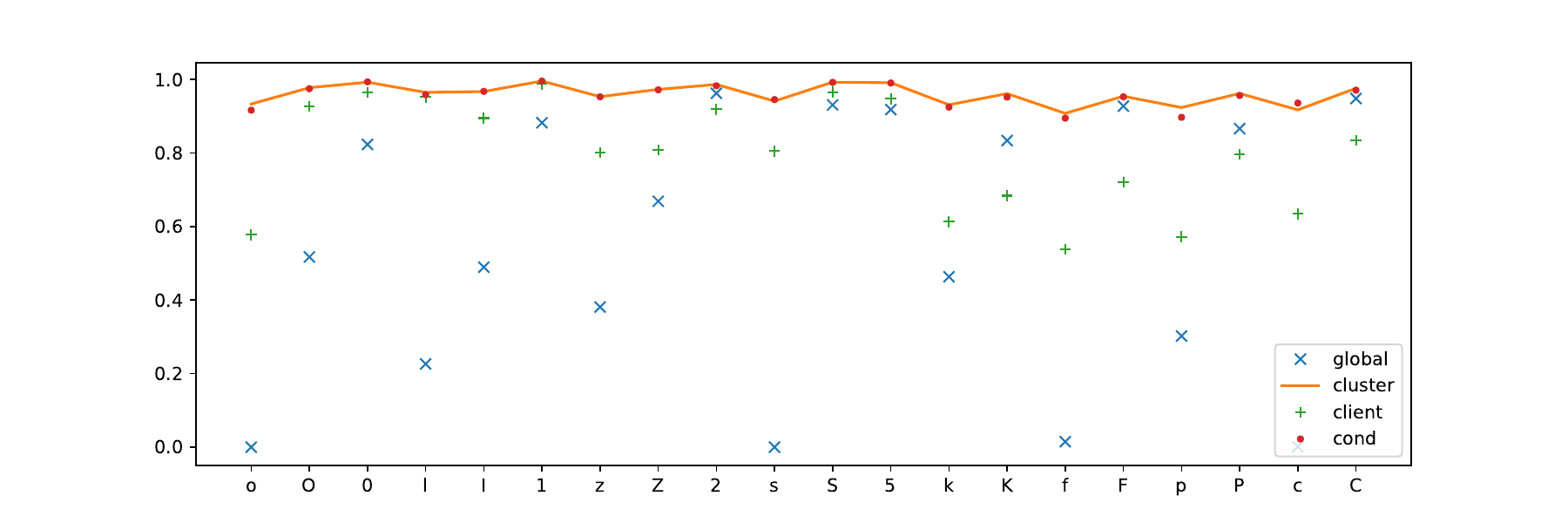}
\caption[MCDC EMNIST: confusable character pairs]{MCDC EMNIST: accuracy on confusable character pairs across subsets. The global model (rightmost) shows systematic failure on lowercase characters with cross-subset visual ambiguity, while the conditional model matches the cluster oracle.}
\label{fig:mcdc_confusable}
\end{figure}

\subsection{Connection to the Present Work}

The FLICS 2026 paper extends the MCDC results in several directions:
\begin{enumerate}
    \item \textbf{PCA eigenvalues} instead of eigenvectors, and computed on learned embeddings rather than raw pixels for CIFAR datasets, providing a compact scalar representation of each client's data distribution
    \item \textbf{Concatenation} at the FC layer (same architecture as EMNIST), with three alternative conditioning architectures (conditional linear, ensemble) dropped in favor of this single, simpler approach
    \item \textbf{Image classification} with CNN on four datasets (MNIST, Fashion-MNIST, CIFAR-10, CIFAR-100) instead of synthetic tasks and character recognition
    \item \textbf{Systematic heterogeneity taxonomy}: label shift, covariate shift, concept shift, and combined heterogeneity (97 configurations)
    \item \textbf{Seven baselines} (FedAvg, Gossip, Local, Oracle, IFCA, DAC, Ditto) instead of three reference models (global, cluster, client)
    \item \textbf{Sparsity analysis} showing unique invariance to client data volume
\end{enumerate}

\textbf{Addressing MCDC reviewer feedback.}
The MCDC reviewers requested evaluation on more diverse and complex datasets beyond EMNIST, robustness analysis across data sparsity levels, and confidence intervals.
The experiments reported in the main paper and in Sections~\ref{sec:e1}--\ref{sec:e4b} of this appendix directly address these concerns with four image classification datasets, systematic sparsity sweeps, and the 97-configuration evaluation.

\section{Experimental Setup for Extended Evaluation}
\label{sec:setup}

Building on the preliminary MCDC results, we now describe the extended experimental framework used throughout the remainder of this appendix.
Compared to the MCDC evaluation (3 clusters, 2 tasks, 3 reference models), the experiments below use four image classification datasets, systematically vary heterogeneity type and severity, sweep data sparsity, and compare against seven baselines spanning all major FL paradigms.

\subsection{Methods}

We evaluate eight methods.
Three serve as reference points: \textbf{Local} trains each client independently on its own data (lower bound on what collaboration can improve upon); \textbf{FedAvg} trains a single global model by averaging local SGD updates across all clients (the standard FL baseline); and \textbf{Oracle} trains one model per ground-truth cluster, pooling data within each cluster (upper bound for clustering-based approaches, since it uses perfect cluster assignments).
\textbf{Gossip} implements naive decentralized learning with random pairwise model averaging on a random communication topology, without adaptive aggregation weights.

Four methods represent established personalization strategies:
\begin{itemize}
    \item \textbf{IFCA}~\cite{ghosh2020ifca}: Iterative Federated Clustering Algorithm, which maintains $K$ model hypotheses and assigns each client to the model with lowest training loss.
    \item \textbf{DAC}~\cite{zec2022dac}: Decentralized Adaptive Clustering, where clients average models with neighbors weighted by cosine similarity of model parameters, enabling implicit clustering without central coordination.
    \item \textbf{Ditto}~\cite{li2021ditto}: each client maintains a personalized model that alternates between global averaging and local fine-tuning with a proximal regularizer ($\lambda = 1.0$).
\end{itemize}

Finally, \textbf{Conditional} (our method) trains a single model conditioned on client-specific PCA eigenvalue statistics via concatenation before the fully-connected layers, as described in the main paper.
Both Conditional and Ditto use the standard FedAvg communication pattern (client$\to$server$\to$client model exchanges), but they differ in how personalization is achieved: Ditto maintains \emph{two} models per client---a shared global model and a separate personal model regularized toward it---requiring double the local computation and storage.
Conditional maintains a \emph{single} shared model and achieves personalization through locally-computed statistics (32 scalars) that never leave the client, requiring no additional communication, no cluster assignments, and no peer-to-peer exchanges.
IFCA requires the server to broadcast $K$ cluster models per round (increasing bandwidth by a factor of $K$), while DAC operates in a fully decentralized topology with peer-to-peer model exchanges.

\subsection{Datasets}

We use four image classification datasets of increasing complexity:
\begin{itemize}
    \item \textbf{MNIST}: 60{,}000 training / 10{,}000 test images of handwritten digits (28$\times$28 grayscale, 10 classes)
    \item \textbf{Fashion-MNIST}: 60{,}000 / 10{,}000 images of clothing items (28$\times$28 grayscale, 10 classes)
    \item \textbf{CIFAR-10}: 50{,}000 / 10{,}000 natural images (32$\times$32 RGB, 10 classes)
    \item \textbf{CIFAR-100}: 50{,}000 / 10{,}000 natural images (32$\times$32 RGB, 100 fine-grained classes grouped into 20 superclasses)
\end{itemize}

For each experiment, clients within a cluster share the same data distribution but receive disjoint subsets of the cluster's data.
Test data is split identically: each client is evaluated on test samples matching its own distribution (local test sets).

\subsection{Client Statistics}
\label{sec:setup_stats}

For the Conditional method, each client $i$ computes a distributional fingerprint from its local training data.
The client constructs an augmented matrix $\mathbf{Z}_i = [\phi(\mathbf{X}_i),\; \text{onehot}(\mathbf{y}_i)]$ by concatenating features with one-hot encoded labels, then performs PCA and retains the top-32 eigenvalues as the statistics vector $\mathbf{s}_i = (\lambda_i^{(1)}, \ldots, \lambda_i^{(32)}) \in \mathbb{R}^{32}$.
The inclusion of labels ensures that the eigenvalues capture label composition, feature structure, and their interaction.

The feature representation $\phi$ differs by dataset:
\begin{itemize}
    \item \textbf{MNIST / Fashion-MNIST}: $\phi$ is the identity on flattened pixels, giving $\mathbf{Z}_i \in \mathbb{R}^{n_i \times (784 + C)}$
    \item \textbf{CIFAR-10 / CIFAR-100}: $\phi$ maps images through a cross-trained convolutional encoder to 128-dimensional embeddings, giving $\mathbf{Z}_i \in \mathbb{R}^{n_i \times (128 + C)}$
\end{itemize}

We use 32 PCA components throughout.
The resulting eigenvalue vectors are z-score normalized across clients before being used as conditioning inputs.

The cross-trained encoder is a small ResNet (two residual stages, 32$\to$64$\to$128 channels, global average pooling to 128 dimensions) trained for classification on a \emph{different} dataset than the target task: for CIFAR-10 experiments, the encoder is trained on CIFAR-100, and vice versa.
This ensures that the embeddings are learned from data that does not overlap with the federated training set, avoiding information leakage.
The encoder is trained centrally for 50 epochs (Adam, lr${}=0.001$, batch size 128) and distributed to all clients, where it is frozen and used solely to produce per-sample embeddings from which the \emph{aggregate} PCA eigenvalue statistics are computed.
Crucially, neither the encoder nor the individual per-sample embeddings enter the training pipeline: only the resulting $k$-dimensional eigenvalue vector (one per client, summarizing the entire local training set) is concatenated as a conditioning input to the federated model.
At test time, the same client-level statistic (computed from training data) is reused; test samples are never passed through the cross-trained encoder.

\subsection{Training Configuration}

All methods use the same CNN backbone (two convolutional layers for 28$\times$28 inputs, four for 32$\times$32 inputs) with identical hyperparameters:
\begin{itemize}
    \item SGD with learning rate 0.01, momentum 0.9
    \item Batch size 64, 20 training epochs
    \item Local test sets matching each client's training distribution
\end{itemize}

For IFCA and DAC, we provide the true number of clusters $K$, giving them an advantage unavailable in practice.
Each configuration uses deterministic data splits; variance across random seeds is small ($< 0.5\%$) so we report single-run results.

\section{E1: Label Shift}
\label{sec:e1}

Label shift heterogeneity occurs when different clients have data from different subsets of classes. We partition classes into $K$ superclass clusters, with each cluster containing a disjoint subset of the original classes.

\subsection{E1: K Sweep Results}

We vary the number of clusters $K \in \{2, 3, 5, 10\}$, with 5 clients per cluster.

\begin{table}[htbp]
\centering
\caption[E1: Label shift accuracy vs K]{E1: Label Shift Accuracy vs Number of Clusters (K)}
\label{tab:e1_k_sweep}
\small
\begin{tabular}{l|cccc|cccc|cccc}
\toprule
& \multicolumn{4}{c|}{\textbf{MNIST}} & \multicolumn{4}{c|}{\textbf{FMNIST}} & \multicolumn{4}{c}{\textbf{CIFAR-10}} \\
Method & K=2 & K=3 & K=5 & K=10 & K=2 & K=3 & K=5 & K=10 & K=2 & K=3 & K=5 & K=10 \\
\midrule
Local       & .991 & .992 & .997 & 1.00 & .908 & .920 & .945 & 1.00 & .767 & .784 & .856 & 1.00 \\
FedAvg      & .986 & .978 & .933 & .522 & .902 & .886 & .745 & .412 & .735 & .554 & .325 & .172 \\
Gossip      & .982 & .983 & .892 & .560 & .895 & .899 & .825 & .560 & .745 & .755 & .533 & .566 \\
Ditto       & .993 & .994 & .996 & 1.00 & .910 & .915 & .933 & .999 & .826 & .822 & .843 & 1.00 \\
IFCA        & .995 & .995 & .999 & 1.00 & .922 & .932 & .945 & 1.00 & .847 & .766 & .906 & 1.00 \\
DAC         & .995 & .995 & .998 & 1.00 & .922 & .931 & .950 & 1.00 & .844 & .859 & .898 & 1.00 \\
\midrule
\rowcolor{oraclecolor!20}
Oracle      & .996 & .996 & .999 & 1.00 & .937 & .945 & .962 & 1.00 & .861 & .886 & .927 & 1.00 \\
\rowcolor{bestcolor!15}
Conditional & \best{.996} & \best{.996} & \best{1.00} & \best{1.00} & \best{.936} & \best{.942} & \best{.964} & \best{1.00} & \best{.860} & \best{.875} & \best{.929} & \best{1.00} \\
\bottomrule
\end{tabular}
\end{table}

\textbf{Key observations:}
\begin{itemize}
    \item FedAvg collapses as $K$ increases: 73.5\% $\to$ 17.2\% on CIFAR-10
    \item Conditional matches Oracle across all $K$ values and datasets
    \item At $K=10$, all personalized methods achieve $\sim$100\% (trivial single-class task)
\end{itemize}

\begin{figure}[htbp]
\centering
\includegraphics[width=\textwidth]{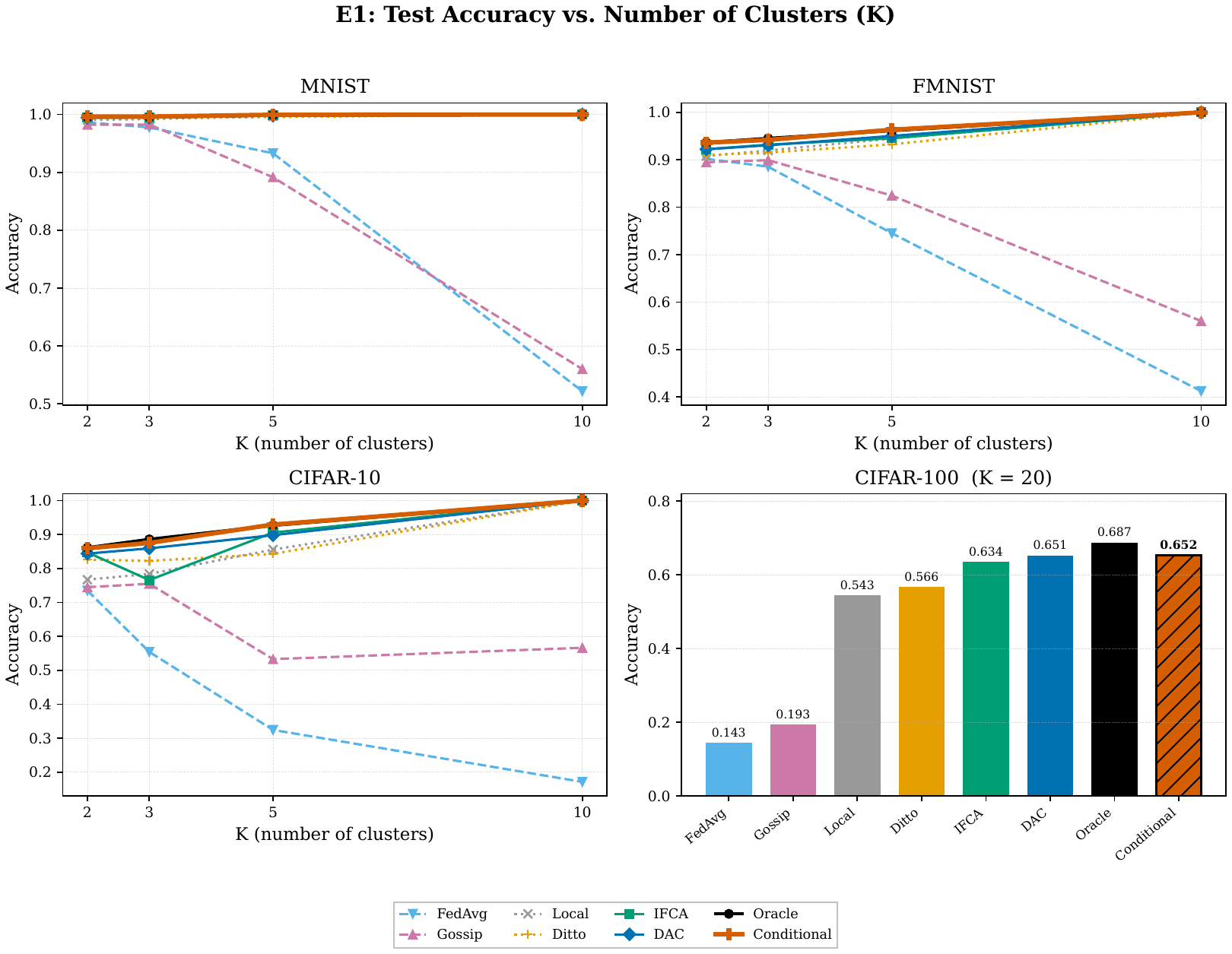}
\caption[E1: Accuracy vs K for label shift]{E1: Test accuracy vs.\ number of clusters (K) for label shift heterogeneity across four datasets. Conditional (our method) matches Oracle performance across all settings.}
\label{fig:e1_accuracy_vs_k}
\end{figure}

\subsection{E1: Sparsity Analysis}

We fix $K=2$ and vary clients per cluster from 5 (Rich, $\sim$6000 samples/client) to 100 (Super Sparse, $\sim$300 samples/client).

\begin{table}[htbp]
\centering
\caption[E1: Label shift accuracy vs sparsity]{E1: Label Shift Accuracy vs Sparsity (K=2, clients per cluster: 5/10/20/50/100)}
\label{tab:e1_sparsity}
\resizebox{\textwidth}{!}{%
\begin{tabular}{l|ccccc|ccccc|ccccc}
\toprule
& \multicolumn{5}{c|}{\textbf{MNIST}} & \multicolumn{5}{c|}{\textbf{FMNIST}} & \multicolumn{5}{c}{\textbf{CIFAR-10}} \\
Method & Rich & Med & Sparse & V.Sp & S.Sp & Rich & Med & Sparse & V.Sp & S.Sp & Rich & Med & Sparse & V.Sp & S.Sp \\
\midrule
Local       & .991 & .986 & .976 & .963 & .940 & .909 & .893 & .866 & .839 & .798 & .767 & .713 & .606 & .526 & .446 \\
FedAvg      & .986 & .980 & .950 & .824 & .738 & .902 & .884 & .837 & .784 & .718 & .709 & .577 & .421 & .472 & .434 \\
Gossip      & .982 & .967 & .873 & .698 & .151 & .895 & .855 & .752 & .633 & .083 & .752 & .694 & .532 & .431 & .211 \\
Ditto       & .993 & .990 & .978 & .948 & .916 & .910 & .891 & .851 & .786 & .743 & .825 & .780 & .695 & .575 & .496 \\
IFCA        & .994 & .992 & .985 & .964 & .935 & .925 & .906 & .872 & .824 & .771 & .852 & .824 & .765 & .715 & .631 \\
DAC         & .994 & .991 & .979 & .957 & .925 & .924 & .902 & .866 & .811 & .753 & .844 & .804 & .714 & .637 & .552 \\
\midrule
\rowcolor{oraclecolor!20}
Oracle      & .996 & .996 & .996 & .996 & .996 & .938 & .938 & .935 & .936 & .936 & .871 & .870 & .871 & .870 & .863 \\
\rowcolor{bestcolor!15}
Conditional & \best{.996} & \best{.996} & \best{.996} & \best{.996} & \best{.997} & \best{.936} & \best{.935} & \best{.935} & \best{.933} & \best{.935} & \best{.862} & \best{.867} & \best{.862} & \best{.860} & \best{.868} \\
\bottomrule
\end{tabular}}
\end{table}

\textbf{Key finding:} Conditional is \textbf{sparsity-invariant}---it maintains Oracle-level performance even at Super Sparse (300 samples/client), while other methods degrade significantly.

\begin{figure}[htbp]
\centering
\includegraphics[width=\textwidth]{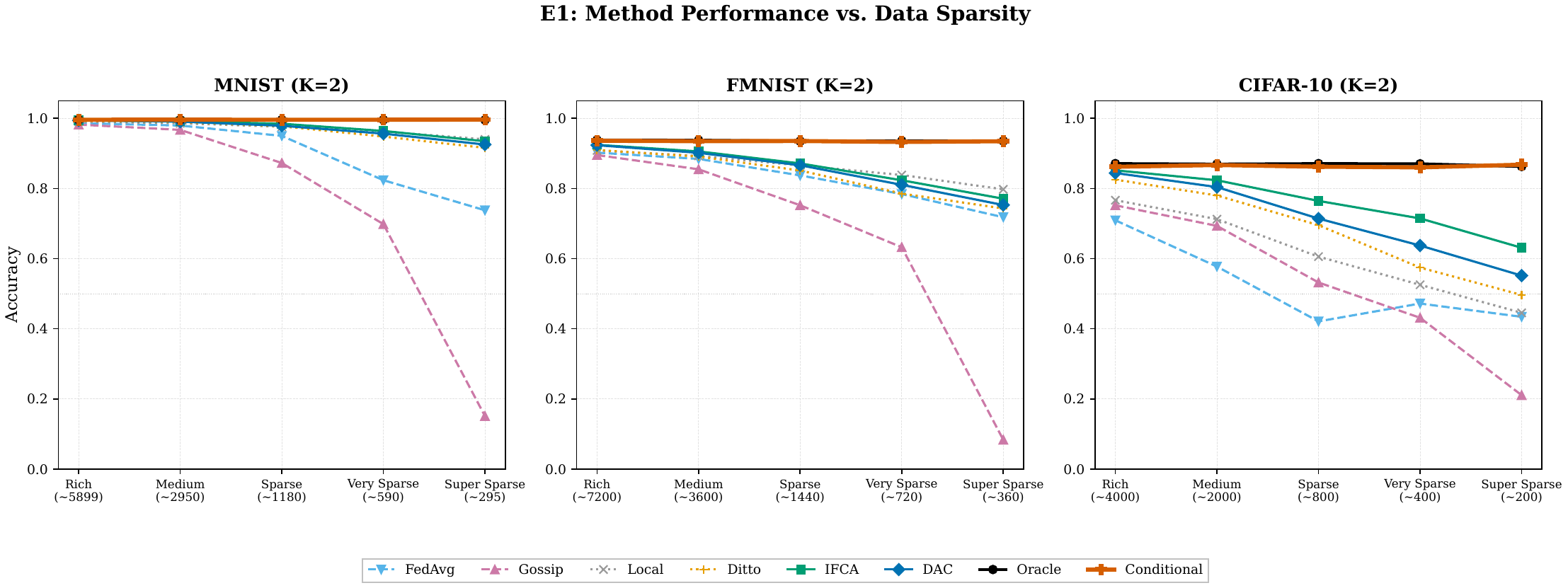}
\caption[E1: Sparsity sweep for label shift]{E1: Method performance vs.\ data sparsity (K=2) for label shift. Conditional maintains Oracle-level accuracy across all sparsity levels.}
\label{fig:e1_sparsity_sweep}
\end{figure}

\subsection{E1: CIFAR-100 (K=20)}

We evaluate on CIFAR-100 using the standard 20-superclass mapping~\cite{krizhevsky2009cifar}, creating 20 clusters with 5 clients each. For example, one superclass groups \emph{apple}, \emph{mushroom}, \emph{orange}, \emph{pear}, and \emph{sweet pepper} under ``fruit and vegetables.''

\begin{table}[htbp]
\centering
\caption[E1: CIFAR-100 label shift (K=20)]{E1: CIFAR-100 Label Shift Accuracy (K=20 superclasses)}
\label{tab:e1_cifar100}
\begin{tabular}{l|c}
\toprule
Method & Accuracy (K=20) \\
\midrule
Local       & .543 \\
FedAvg      & .143 \\
Gossip      & .193 \\
Ditto       & .566 \\
IFCA        & .635 \\
DAC         & .651 \\
\midrule
\rowcolor{oraclecolor!20}
Oracle      & .687 \\
\rowcolor{bestcolor!15}
Conditional & \best{.652} \\
\bottomrule
\end{tabular}
\end{table}

\textbf{Key observation:} CIFAR-100 with 20 superclasses is a much harder task. FedAvg collapses to 14.3\%. Conditional achieves 65.2\%, the best non-oracle result and close to DAC (65.1\%). Oracle reaches 68.7\%.

\begin{figure}[htbp]
\centering
\includegraphics[width=\textwidth]{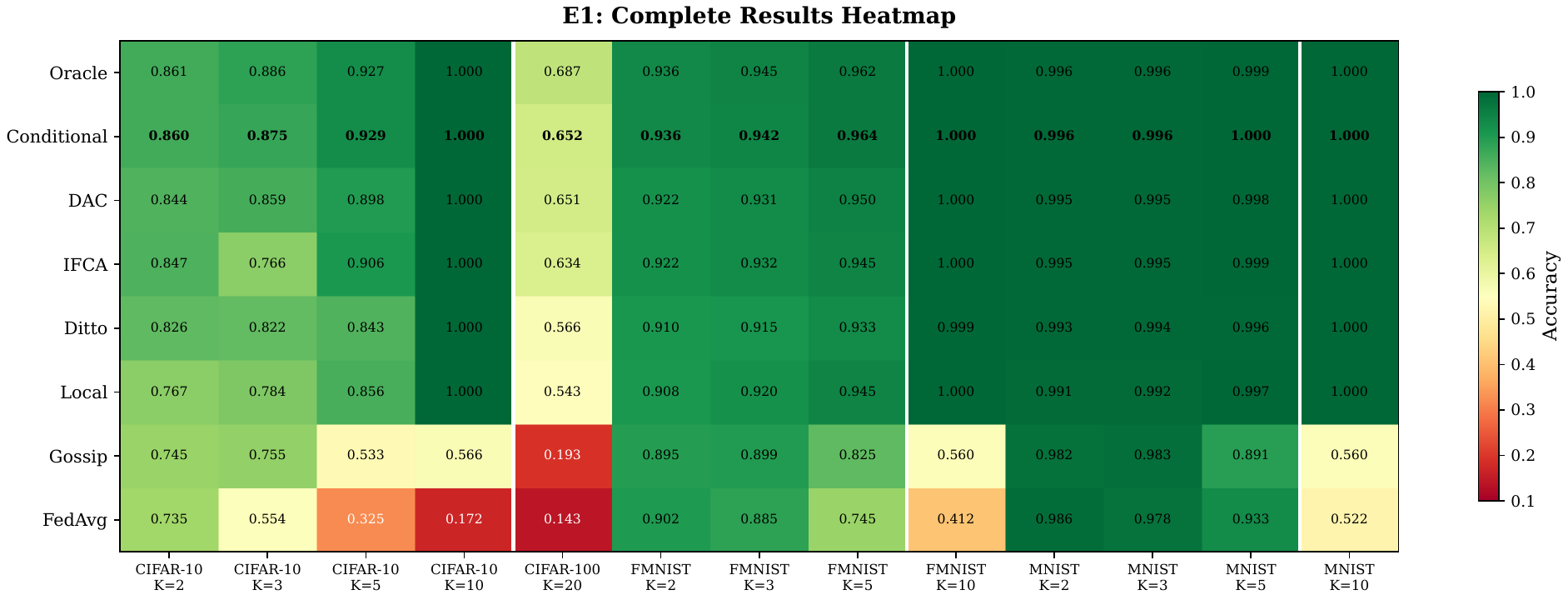}
\caption[E1: Results heatmap]{E1: Complete results heatmap across all datasets and K values. Conditional (our method, bold) consistently matches Oracle performance.}
\label{fig:e1_heatmap}
\end{figure}

\section{E2a: Covariate Shift (Subsampling)}
\label{sec:e2a}

Covariate shift occurs when the input distribution $P(X)$ differs across clients while the labeling function $P(Y|X)$ remains the same. In E2a, different clusters sample from different subclasses within each superclass, creating distinct input distributions with shared label semantics.

\subsection{E2a: K Sweep Results}

We vary $K \in \{2, 3, 4\}$ covariate clusters, with 5 clients per cluster. All clusters predict the same superclass labels but see different fine-grained subclasses. Each client is evaluated on test samples drawn from the same subclass distribution as its training data (local evaluation).

\begin{table}[htbp]
\centering
\caption[E2a: Covariate shift accuracy vs K]{E2a: Covariate Shift (Subsampling) Accuracy vs K}
\label{tab:e2a_k}
\small
\begin{tabular}{l|ccc|ccc|ccc}
\toprule
& \multicolumn{3}{c|}{\textbf{MNIST}} & \multicolumn{3}{c|}{\textbf{FMNIST}} & \multicolumn{3}{c}{\textbf{CIFAR-10}} \\
Method & K=2 & K=3 & K=4 & K=2 & K=3 & K=4 & K=2 & K=3 & K=4 \\
\midrule
Local       & .995 & .996 & .997 & .996 & .996 & .998 & .944 & .955 & .948 \\
FedAvg      & .989 & .981 & .981 & .992 & .986 & .980 & .956 & .943 & .923 \\
Gossip      & .988 & .985 & .980 & .989 & .988 & .989 & .946 & .952 & .932 \\
Ditto       & .996 & .996 & .996 & .995 & .993 & .997 & .960 & .969 & .961 \\
IFCA        & .997 & .996 & .989 & .992 & .996 & .999 & .954 & .943 & .961 \\
DAC         & .997 & .997 & .998 & .997 & .996 & .999 & .967 & .972 & .969 \\
\midrule
\rowcolor{oraclecolor!20}
Oracle      & .999 & .998 & 1.00 & .998 & .997 & .999 & .976 & .979 & .978 \\
\rowcolor{bestcolor!15}
Conditional & \best{.998} & \best{.999} & \best{.999} & \best{.998} & \best{.998} & \best{.999} & \best{.976} & \best{.981} & \best{.982} \\
\bottomrule
\end{tabular}
\end{table}

\textbf{Key observations:}
\begin{itemize}
    \item With local test evaluation, all methods achieve high accuracy on MNIST/FMNIST ($>$98\%)
    \item On CIFAR-10, Conditional matches or exceeds Oracle across all K values
    \item FedAvg remains competitive under pure covariate shift since all clusters share the same label semantics
\end{itemize}

\begin{figure}[htbp]
\centering
\includegraphics[width=\textwidth]{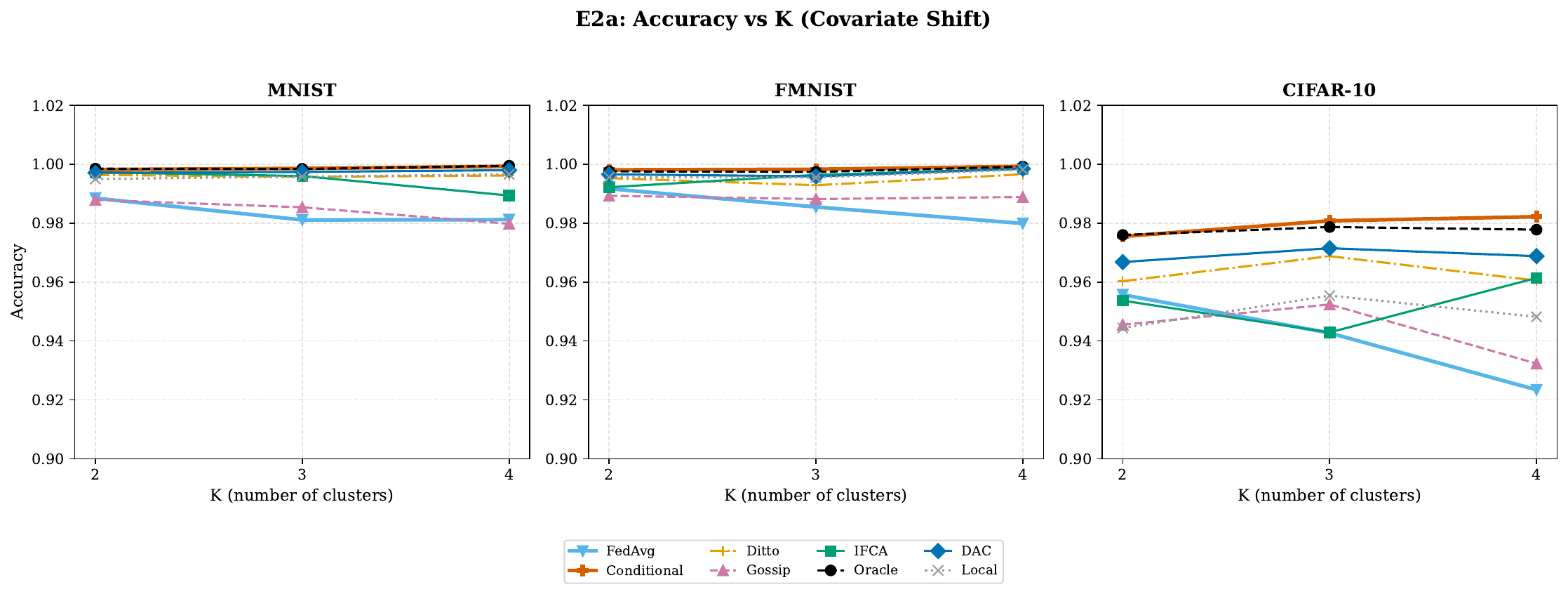}
\caption[E2a: Accuracy vs K for covariate shift]{E2a: Accuracy vs.\ K for covariate shift (subsampling). With local test evaluation, Conditional matches Oracle across all datasets.}
\label{fig:e2a_accuracy_vs_k}
\end{figure}

\subsection{E2a: Sparsity Analysis}

We fix $K=2$ and vary clients per cluster from 5 (Rich) to 100 (Super Sparse).

\begin{table}[htbp]
\centering
\caption[E2a: Covariate shift sparsity sweep]{E2a: Covariate Shift Sparsity Sweep (K=2, clients per cluster: 5/10/20/50/100)}
\label{tab:e2a_sparsity}
\resizebox{\textwidth}{!}{%
\begin{tabular}{l|ccccc|ccccc|ccccc}
\toprule
& \multicolumn{5}{c|}{\textbf{MNIST}} & \multicolumn{5}{c|}{\textbf{FMNIST}} & \multicolumn{5}{c}{\textbf{CIFAR-10}} \\
Method & Rich & Med & Sparse & V.Sp & S.Sp & Rich & Med & Sparse & V.Sp & S.Sp & Rich & Med & Sparse & V.Sp & S.Sp \\
\midrule
Local       & .995 & .992 & .987 & .980 & .967 & .995 & .994 & .992 & .990 & .989 & .946 & .926 & .906 & .881 & .863 \\
FedAvg      & .988 & .986 & .977 & .932 & .847 & .992 & .988 & .986 & .985 & .984 & .956 & .945 & .932 & .924 & .916 \\
Gossip      & .988 & .981 & .897 & .505 & .504 & .989 & .988 & .987 & .600 & .600 & .951 & .933 & .907 & .854 & .833 \\
Ditto       & .997 & .995 & .989 & .968 & .928 & .995 & .993 & .991 & .990 & .988 & .961 & .946 & .936 & .915 & .892 \\
IFCA        & .998 & .996 & .991 & .981 & .958 & .992 & .989 & .986 & .986 & .984 & .954 & .948 & .933 & .922 & .915 \\
DAC         & .997 & .995 & .990 & .977 & .955 & .997 & .994 & .992 & .991 & .990 & .966 & .954 & .925 & .920 & .900 \\
\midrule
\rowcolor{oraclecolor!20}
Oracle      & .999 & .999 & .998 & .998 & .999 & .998 & .998 & .998 & .998 & .998 & .975 & .973 & .977 & .971 & .974 \\
\rowcolor{bestcolor!15}
Conditional & \best{.998} & \best{.998} & \best{.998} & \best{.998} & \best{.998} & \best{.998} & \best{.998} & \best{.998} & \best{.998} & \best{.998} & \best{.974} & \best{.972} & \best{.974} & \best{.973} & \best{.972} \\
\bottomrule
\end{tabular}}
\end{table}

\textbf{Key finding:} Conditional is perfectly sparsity-invariant under covariate shift, maintaining $\sim$99.8\% on MNIST/FMNIST and $\sim$97.3\% on CIFAR-10 across all sparsity levels. Gossip collapses to 50--60\% at very sparse settings.

\begin{figure}[htbp]
\centering
\includegraphics[width=\textwidth]{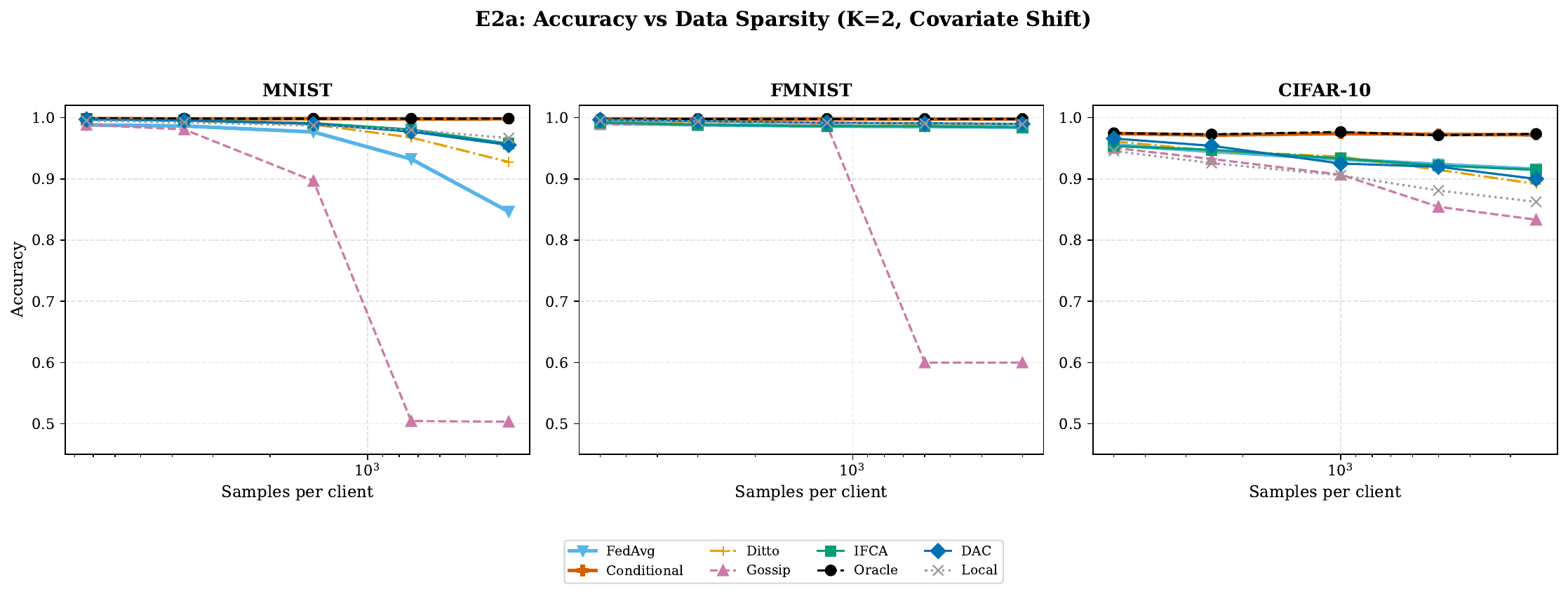}
\caption[E2a: Sparsity sweep for covariate shift]{E2a: Accuracy vs.\ data sparsity (K=2) for covariate shift. Conditional is perfectly sparsity-invariant, while Gossip collapses at sparse settings.}
\label{fig:e2a_sparsity_sweep}
\end{figure}

\section{E2b: Covariate Shift (Rotation)}
\label{sec:e2b}

For MNIST and FMNIST, we create covariate shift by applying different rotation angles to each cluster, following the rotation-based heterogeneity setup used in decentralized learning evaluations~\cite{zec2022dac}.
We test four rotation configurations of increasing complexity:
\begin{itemize}
    \item \textbf{C1}: $K=2$ --- $0\degree$, $180\degree$
    \item \textbf{C2}: $K=3$ --- $0\degree$, $120\degree$, $240\degree$
    \item \textbf{C3}: $K=4$ --- $0\degree$, $90\degree$, $180\degree$, $270\degree$
    \item \textbf{C4}: $K=4$ --- $0\degree$, $10\degree$, $180\degree$, $350\degree$ (subtle + major)
\end{itemize}

\begin{table}[htbp]
\centering
\caption[E2b: Rotation covariate shift]{E2b: Rotation-Based Covariate Shift across four configurations}
\label{tab:e2b}
\small
\begin{tabular}{l|cccc|cccc}
\toprule
& \multicolumn{4}{c|}{\textbf{MNIST}} & \multicolumn{4}{c}{\textbf{FMNIST}} \\
Method & C1 & C2 & C3 & C4 & C1 & C2 & C3 & C4 \\
\midrule
Local       & .981 & .976 & .972 & .971 & .877 & .860 & .852 & .851 \\
FedAvg      & .976 & .962 & .940 & .965 & .876 & .836 & .789 & .821 \\
Gossip      & .968 & .965 & .931 & .949 & .859 & .810 & .718 & .773 \\
Ditto       & .985 & .976 & .971 & .979 & .880 & .846 & .823 & .837 \\
IFCA        & .990 & .981 & .977 & .983 & .880 & .872 & .860 & .855 \\
DAC         & .989 & .985 & .982 & .982 & .893 & .871 & .868 & .863 \\
\midrule
\rowcolor{oraclecolor!20}
Oracle      & .992 & .989 & .987 & .987 & .912 & .903 & .900 & .899 \\
\rowcolor{bestcolor!15}
Conditional & \best{.990} & \best{.986} & \best{.986} & \best{.988} & \best{.913} & \best{.898} & \best{.895} & \best{.903} \\
\bottomrule
\end{tabular}
\end{table}

\textbf{Key observations:}
\begin{itemize}
    \item Conditional matches Oracle within 0.3\% on MNIST and within 0.5\% on FMNIST across all configurations
    \item On FMNIST C1 and C4, Conditional slightly exceeds Oracle ($+0.1\%$, $+0.4\%$)
    \item FedAvg degrades significantly with more rotation diversity (MNIST C3: 94.0\%, FMNIST C3: 78.9\%)
    \item PCA eigenvalue statistics effectively capture rotation characteristics
\end{itemize}

\begin{figure}[htbp]
\centering
\includegraphics[width=\textwidth]{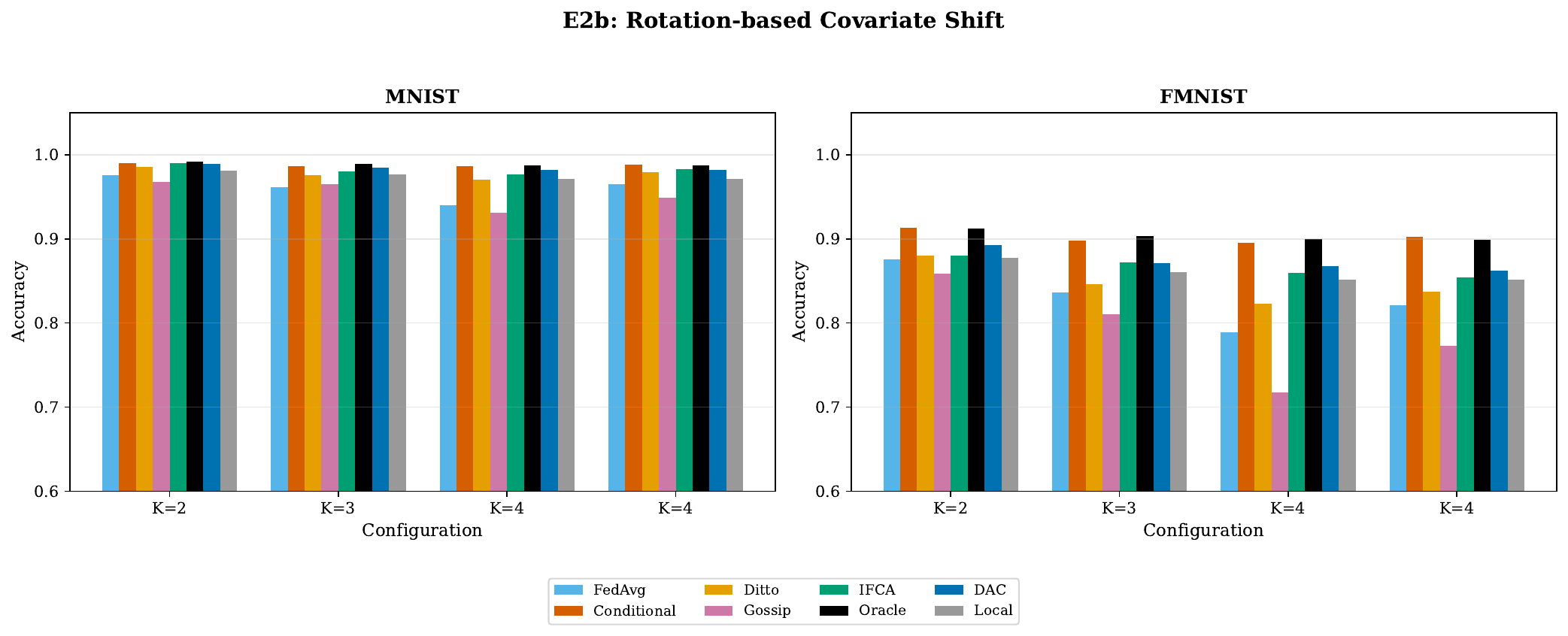}
\caption[E2b: Rotation covariate shift results]{E2b: Rotation-based covariate shift results across all four configurations. Conditional closely tracks Oracle across all rotation settings.}
\label{fig:e2b_rotation}
\end{figure}

\section{E3a: Semantic Concept Shift}
\label{sec:e3a}

Concept shift occurs when the relationship between inputs and labels differs across clients---the same image may have different ``correct'' labels depending on the client's task definition.

In E3a, different clusters use different semantic groupings for the same images:
\begin{itemize}
    \item \textbf{MNIST}: odd/even vs below5/above5
    \item \textbf{FMNIST}: formal/casual vs seasonal (summer/winter)
    \item \textbf{CIFAR-10}: semantic (vehicle/domestic/wild) vs size (large/medium/small)
\end{itemize}

\begin{table}[htbp]
\centering
\caption[E3a: Semantic concept shift sparsity sweep]{E3a: Semantic Concept Shift Accuracy (Sparsity Sweep, K=2, clients per cluster: 5/10/20/50/100)}
\label{tab:e3a}
\resizebox{\textwidth}{!}{%
\begin{tabular}{l|ccccc|ccccc|ccccc}
\toprule
& \multicolumn{5}{c|}{\textbf{MNIST}} & \multicolumn{5}{c|}{\textbf{FMNIST}} & \multicolumn{5}{c}{\textbf{CIFAR-10}} \\
Method & Rich & Med & Sparse & V.Sp & S.Sp & Rich & Med & Sparse & V.Sp & S.Sp & Rich & Med & Sparse & V.Sp & S.Sp \\
\midrule
Local       & .985 & .979 & .966 & .948 & .915 & .930 & .919 & .901 & .880 & .854 & .829 & .774 & .741 & .707 & .676 \\
FedAvg      & .793 & .791 & .785 & .768 & .713 & .671 & .661 & .639 & .628 & .636 & .762 & .746 & .742 & .733 & .727 \\
Gossip      & .871 & .866 & .506 & .502 & .500 & .774 & .746 & .535 & .500 & .500 & .804 & .771 & .716 & .693 & .649 \\
Ditto       & .988 & .983 & .963 & .910 & .765 & .928 & .918 & .868 & .815 & .697 & .878 & .837 & .804 & .756 & .718 \\
IFCA        & .992 & .988 & .962 & .767 & .714 & .943 & .934 & .905 & .876 & .798 & .759 & .747 & .740 & .734 & .726 \\
DAC         & .992 & .987 & .974 & .944 & .829 & .941 & .931 & .902 & .869 & .816 & .847 & .844 & .806 & .758 & .735 \\
\midrule
\rowcolor{oraclecolor!20}
Oracle      & .993 & .994 & .993 & .994 & .994 & .950 & .949 & .949 & .950 & .948 & .893 & .891 & .898 & .899 & .899 \\
\rowcolor{bestcolor!15}
Conditional & \best{.994} & \best{.994} & \best{.994} & \best{.992} & \best{.991} & \best{.950} & \best{.951} & \best{.947} & \best{.943} & \best{.940} & \best{.904} & \best{.891} & \best{.884} & \best{.870} & \best{.814} \\
\bottomrule
\end{tabular}}
\end{table}

\textbf{Key observations:}
\begin{itemize}
    \item Conditional matches Oracle on MNIST/FMNIST and exceeds it on CIFAR-10 Rich ($+1.1\%$)
    \item FedAvg averages conflicting semantic groupings, achieving only 63--79\%
    \item Gossip collapses to 50\% (random) on MNIST and FMNIST at sparse settings
    \item IFCA struggles with cluster estimation at very sparse settings
    \item On CIFAR-10, Conditional shows some degradation at very sparse settings (90.4\% Rich $\to$ 81.4\% Super Sparse), narrowing the gap to Oracle (89.9\%).
    However, it still outperforms DAC (73.5\%), IFCA (72.6\%), and Ditto (71.8\%) by a substantial margin at Super Sparse.
    The degradation likely reflects that with very few samples per client, the PCA eigenvalue estimates become noisier, providing less precise conditioning signals.
\end{itemize}

\begin{figure}[htbp]
\centering
\includegraphics[width=\textwidth]{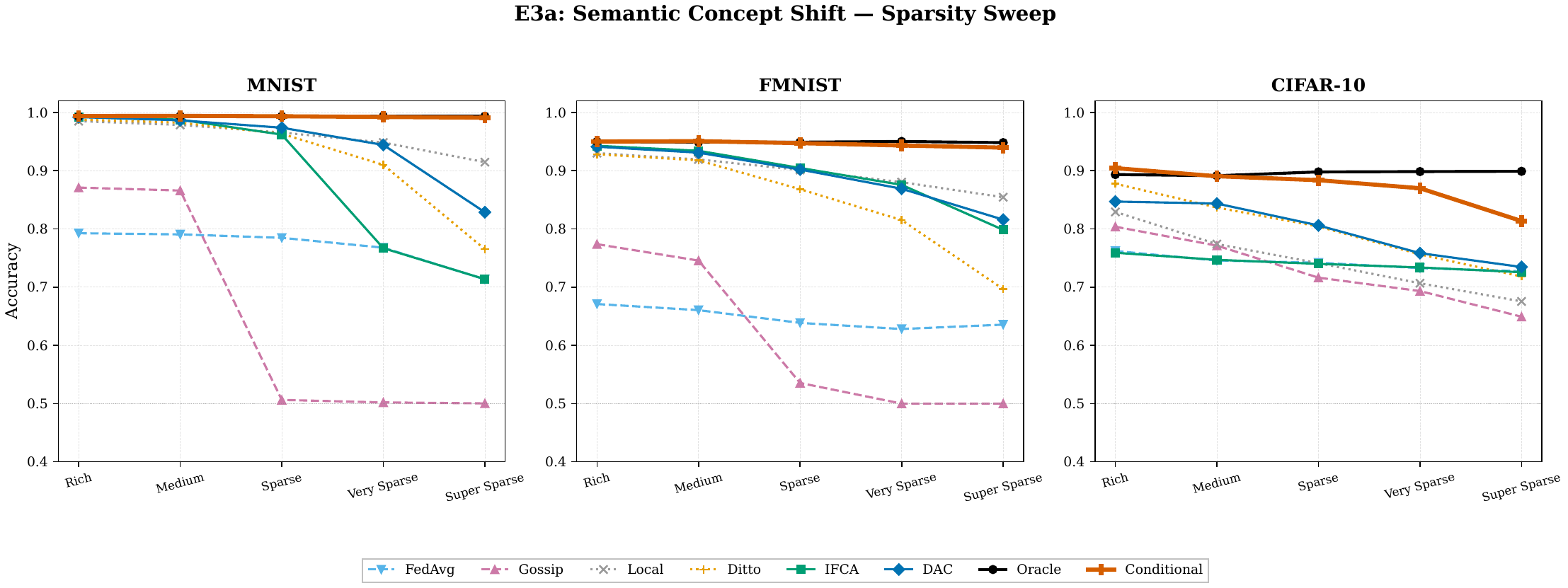}
\caption[E3a: Semantic concept shift sparsity sweep]{E3a: Semantic concept shift sparsity sweep. Conditional maintains near-Oracle performance while other methods degrade significantly.}
\label{fig:e3a_sparsity}
\end{figure}

\begin{figure}[htbp]
\centering
\includegraphics[width=\textwidth]{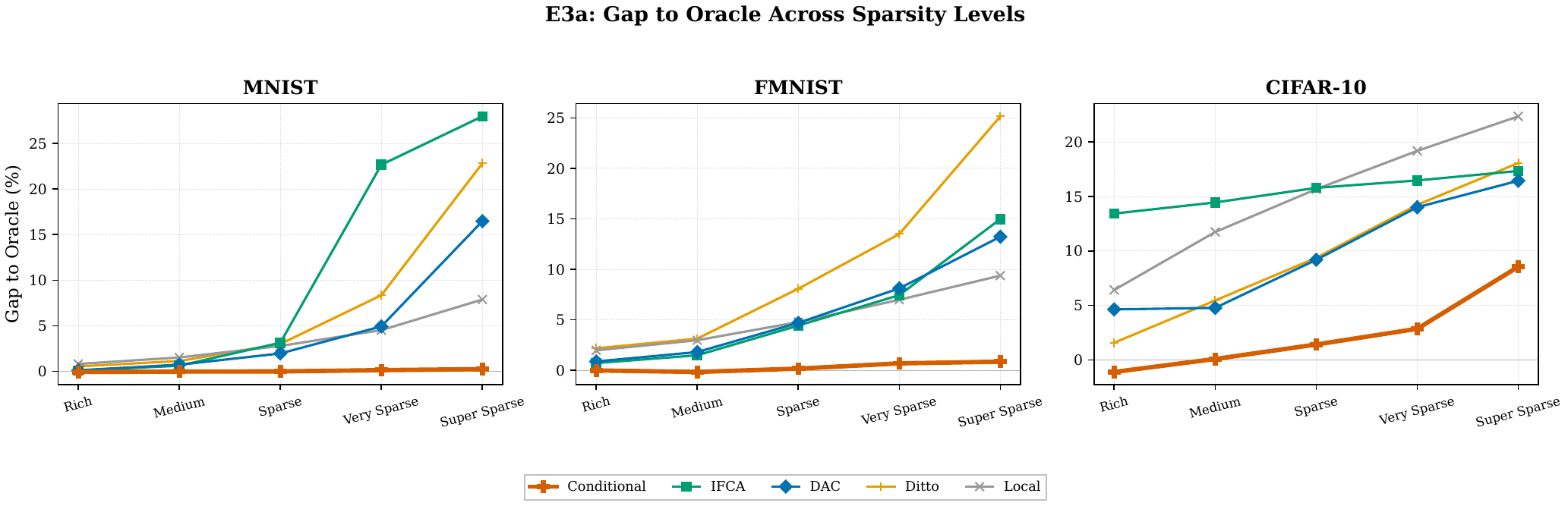}
\caption[E3a: Gap to Oracle vs sparsity]{E3a: Gap to Oracle across sparsity levels (personalized methods only). Conditional stays closest to Oracle (negative gap = better than Oracle).}
\label{fig:e3a_oracle_gap}
\end{figure}

\section{E3b: Label Permutation}
\label{sec:e3b}

Each cluster uses a different random permutation of the 10-class labels, creating maximal concept shift---the same image receives completely different labels across clusters.

\subsection{E3b: K Sweep}

We vary $K \in \{2, 3, 4, 5\}$ permutation clusters with 5 clients per cluster.

\begin{table}[htbp]
\centering
\caption[E3b: Label permutation K sweep]{E3b: Label Permutation --- K Sweep}
\label{tab:e3b_k}
\small
\begin{tabular}{l|cccc|cccc|cccc}
\toprule
& \multicolumn{4}{c|}{\textbf{MNIST}} & \multicolumn{4}{c|}{\textbf{FMNIST}} & \multicolumn{4}{c}{\textbf{CIFAR-10}} \\
Method & K=2 & K=3 & K=4 & K=5 & K=2 & K=3 & K=4 & K=5 & K=2 & K=3 & K=4 & K=5 \\
\midrule
Local       & .981 & .977 & .973 & .969 & .877 & .863 & .854 & .845 & .668 & .618 & .579 & .558 \\
FedAvg      & .495 & .332 & .295 & .306 & .454 & .330 & .278 & .257 & .408 & .304 & .285 & .282 \\
Gossip      & .691 & .798 & .568 & .585 & .619 & .677 & .477 & .484 & .545 & .583 & .429 & .401 \\
Ditto       & .982 & .980 & .975 & .970 & .877 & .854 & .833 & .815 & .767 & .725 & .685 & .661 \\
IFCA        & .990 & .985 & .982 & .844 & .892 & .593 & .659 & .705 & .411 & .535 & .717 & .574 \\
DAC         & .989 & .986 & .982 & .979 & .891 & .878 & .866 & .856 & .773 & .743 & .709 & .693 \\
\midrule
\rowcolor{oraclecolor!20}
Oracle      & .992 & .988 & .988 & .986 & .914 & .904 & .899 & .894 & .795 & .767 & .743 & .728 \\
\rowcolor{bestcolor!15}
Conditional & \best{.992} & \best{.991} & \best{.991} & \best{.989} & \best{.916} & \best{.914} & \best{.911} & \best{.910} & \best{.808} & \best{.804} & \best{.790} & \best{.788} \\
\bottomrule
\end{tabular}
\end{table}

\textbf{Key observations:}
\begin{itemize}
    \item FedAvg collapses to near-random ($\sim$30\%) as label conflicts destroy information
    \item \textbf{Conditional beats Oracle by 1--6\% on CIFAR-10}---statistics encode more useful information than cluster ID alone
    \item IFCA shows inconsistent behavior, sometimes achieving ARI=1.0, sometimes ARI=0.0
\end{itemize}

\begin{figure}[htbp]
\centering
\includegraphics[width=\textwidth]{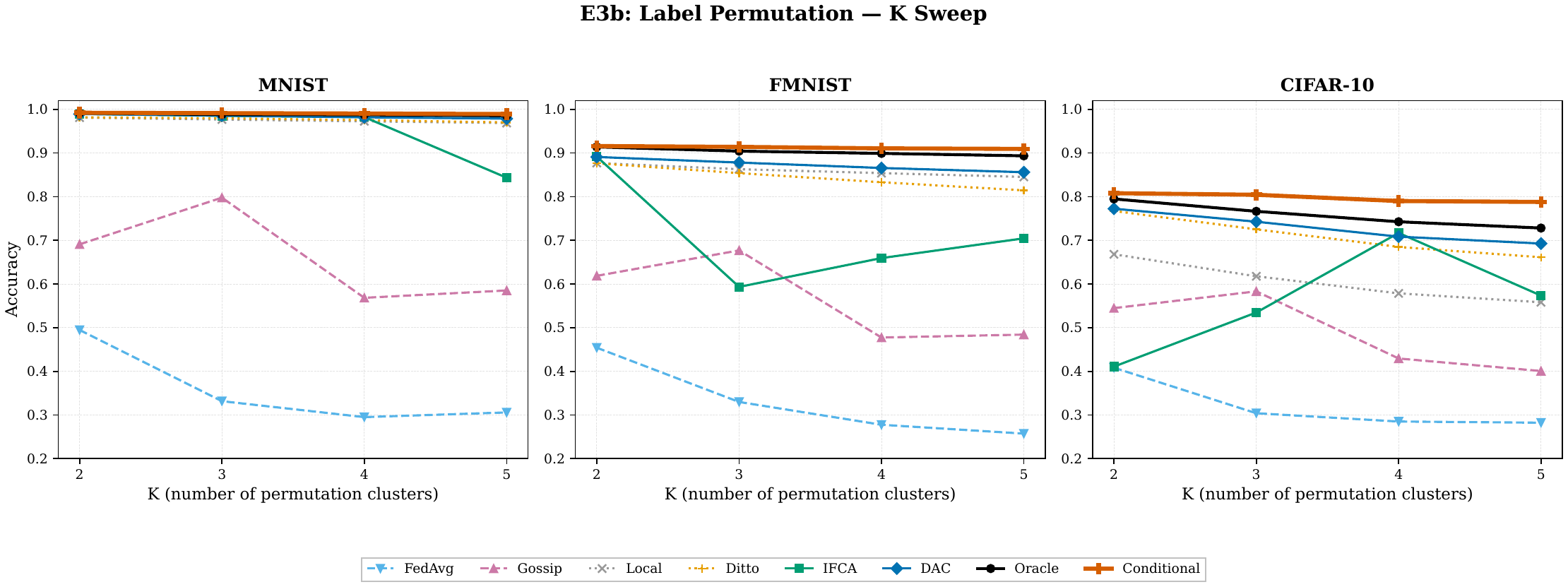}
\caption[E3b: Label permutation K sweep]{E3b: Label permutation K sweep. Conditional consistently outperforms Oracle on CIFAR-10, demonstrating the value of continuous statistics over discrete cluster assignments.}
\label{fig:e3b_k_sweep}
\end{figure}

\subsection{E3b: Sparsity Analysis}

We fix $K=2$ and vary clients per cluster from 5 (Rich) to 100 (Super Sparse).

\begin{table}[htbp]
\centering
\caption[E3b: Label permutation sparsity sweep]{E3b: Label Permutation --- Sparsity Sweep (K=2, clients per cluster: 5/10/20/50/100)}
\label{tab:e3b_sparsity}
\resizebox{\textwidth}{!}{%
\begin{tabular}{l|ccccc|ccccc|ccccc}
\toprule
& \multicolumn{5}{c|}{\textbf{MNIST}} & \multicolumn{5}{c|}{\textbf{FMNIST}} & \multicolumn{5}{c}{\textbf{CIFAR-10}} \\
Method & Rich & Med & Sparse & V.Sp & S.Sp & Rich & Med & Sparse & V.Sp & S.Sp & Rich & Med & Sparse & V.Sp & S.Sp \\
\midrule
Local       & .981 & .973 & .954 & .929 & .880 & .878 & .854 & .814 & .768 & .715 & .668 & .576 & .467 & .406 & .349 \\
FedAvg      & .494 & .492 & .481 & .461 & .399 & .456 & .440 & .419 & .388 & .367 & .411 & .391 & .354 & .322 & .286 \\
Gossip      & .691 & .683 & .107 & .107 & .106 & .621 & .584 & .493 & .100 & .100 & .544 & .497 & .425 & .307 & .207 \\
Ditto       & .983 & .976 & .949 & .897 & .766 & .877 & .848 & .781 & .737 & .663 & .769 & .719 & .612 & .494 & .394 \\
IFCA        & .990 & .986 & .970 & .936 & .869 & .892 & .873 & .816 & .774 & .721 & .412 & .739 & .675 & .591 & .502 \\
DAC         & .989 & .982 & .963 & .920 & .849 & .889 & .865 & .805 & .756 & .702 & .774 & .712 & .606 & .522 & .451 \\
\midrule
\rowcolor{oraclecolor!20}
Oracle      & .992 & .991 & .992 & .992 & .991 & .913 & .915 & .913 & .914 & .914 & .796 & .792 & .792 & .792 & .791 \\
\rowcolor{bestcolor!15}
Conditional & \best{.992} & \best{.992} & \best{.992} & \best{.991} & \best{.987} & \best{.918} & \best{.919} & \best{.914} & \best{.909} & \best{.901} & \best{.806} & \best{.816} & \best{.795} & \best{.772} & \best{.688} \\
\bottomrule
\end{tabular}}
\end{table}

\textbf{Key observations:}
\begin{itemize}
    \item Conditional maintains near-Oracle performance on MNIST/FMNIST even at Super Sparse
    \item On CIFAR-10, Conditional beats Oracle at Rich and Medium but degrades at Super Sparse (68.8\% vs Oracle 79.1\%)
    \item Gossip collapses to 10\% (random for 10-class) on MNIST at sparse settings and FMNIST at very sparse
    \item FedAvg stays near 30--50\% across all sparsity levels (conflicting labels are equally destructive regardless of data volume)
\end{itemize}

\begin{figure}[htbp]
\centering
\includegraphics[width=\textwidth]{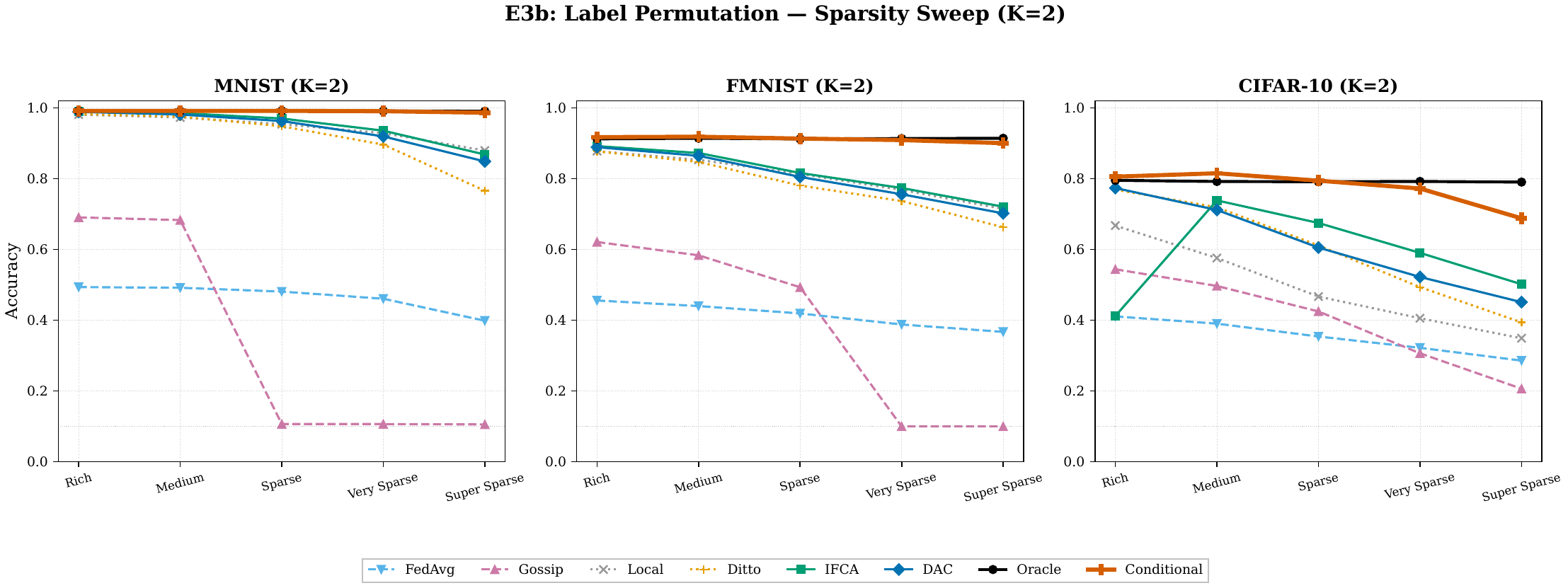}
\caption[E3b: Label permutation sparsity sweep]{E3b: Label permutation sparsity sweep (K=2). Gossip collapses to random at sparse settings; Conditional remains the best practical method.}
\label{fig:e3b_sparsity}
\end{figure}

\section{E4a: Cross-Dataset Domain Shift}
\label{sec:e4a}

Domain shift combines multiple sources of heterogeneity or uses fundamentally different data domains.

E4a uses two clusters from completely different datasets (same 28$\times$28 resolution):
\begin{itemize}
    \item Cluster 0: MNIST (handwritten digits)
    \item Cluster 1: Fashion-MNIST (clothing items)
\end{itemize}

\begin{table}[htbp]
\centering
\caption[E4a: MNIST+FMNIST domain shift sparsity]{E4a: MNIST + FMNIST Domain Shift (Sparsity Sweep)}
\label{tab:e4a}
\begin{tabular}{l|cccc}
\toprule
Method & Rich & Medium & Sparse & Very Sparse \\
& (10 clients) & (20 clients) & (50 clients) & (100 clients) \\
\midrule
Local       & .942 & .928 & .908 & .885 \\
FedAvg      & .939 & .928 & .904 & .865 \\
Gossip      & .936 & .920 & .861 & \textcolor{red}{.109} \\
Ditto       & .944 & .933 & .910 & .872 \\
IFCA        & .953 & .944 & .926 & .896 \\
DAC         & .952 & .941 & .919 & .886 \\
\midrule
\rowcolor{oraclecolor!20}
Oracle      & .957 & .957 & .958 & .959 \\
\rowcolor{bestcolor!15}
Conditional & \best{.955} & \best{.957} & \best{.956} & \best{.959} \\
\bottomrule
\end{tabular}
\end{table}

\textbf{Key observations:}
\begin{itemize}
    \item Conditional matches Oracle perfectly---sparsity-invariant (95.5\% $\to$ 95.9\%)
    \item IFCA achieves perfect clustering (ARI=1.0) across all sparsity levels---the domain difference is trivially detectable
    \item Gossip collapses to 10.9\% at Very Sparse
\end{itemize}

\begin{figure}[htbp]
\centering
\includegraphics[width=0.85\textwidth]{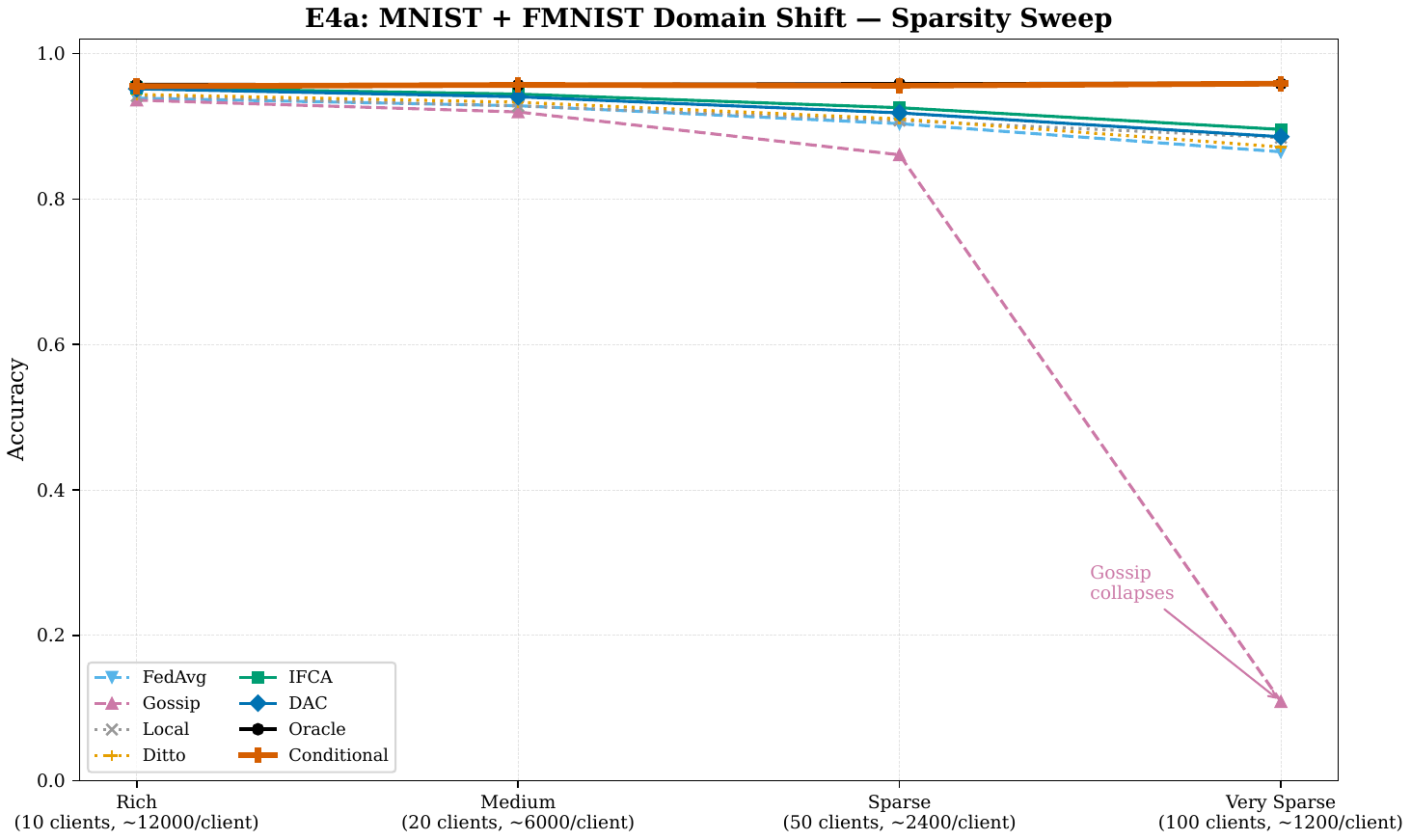}
\caption[E4a: Domain shift sparsity sweep]{E4a: MNIST + FMNIST domain shift sparsity sweep. Conditional matches Oracle; Gossip collapses at Very Sparse.}
\label{fig:e4a_sparsity}
\end{figure}

\section{E4b: Combined Heterogeneity}
\label{sec:e4b}

E4b stacks concept shift (E3a) with covariate shift (E2a) on CIFAR-10:
\begin{itemize}
    \item 2 concept groups (semantic vs size mappings)
    \item $C$ covariate clusters per concept group
    \item Total clusters = $2 \times C$
\end{itemize}

\subsection{E4b: Configuration Sweep}

\begin{table}[htbp]
\centering
\caption[E4b: Combined heterogeneity config sweep]{E4b: Combined Heterogeneity --- Configuration Sweep (CIFAR-10)}
\label{tab:e4b_config}
\begin{tabular}{l|ccc}
\toprule
Method & C=2 (4 clusters) & C=3 (6 clusters) & C=4 (8 clusters) \\
\midrule
Local       & .818 & .854 & .886 \\
FedAvg      & .754 & .743 & .764 \\
Gossip      & .775 & .789 & .786 \\
Ditto       & .868 & .880 & .906 \\
IFCA        & .752 & .740 & .892 \\
DAC         & .888 & .854 & .896 \\
\midrule
\rowcolor{oraclecolor!20}
Oracle      & .891 & .914 & .939 \\
\rowcolor{bestcolor!15}
Conditional & \best{.920} & \best{.935} & \best{.955} \\
\bottomrule
\end{tabular}
\end{table}

\textbf{Key finding:} \textbf{Conditional beats Oracle by 1.5--3\%} on combined heterogeneity. The PCA statistics capture both concept AND covariate dimensions, providing richer conditioning than cluster membership alone.

\begin{figure}[htbp]
\centering
\includegraphics[width=0.85\textwidth]{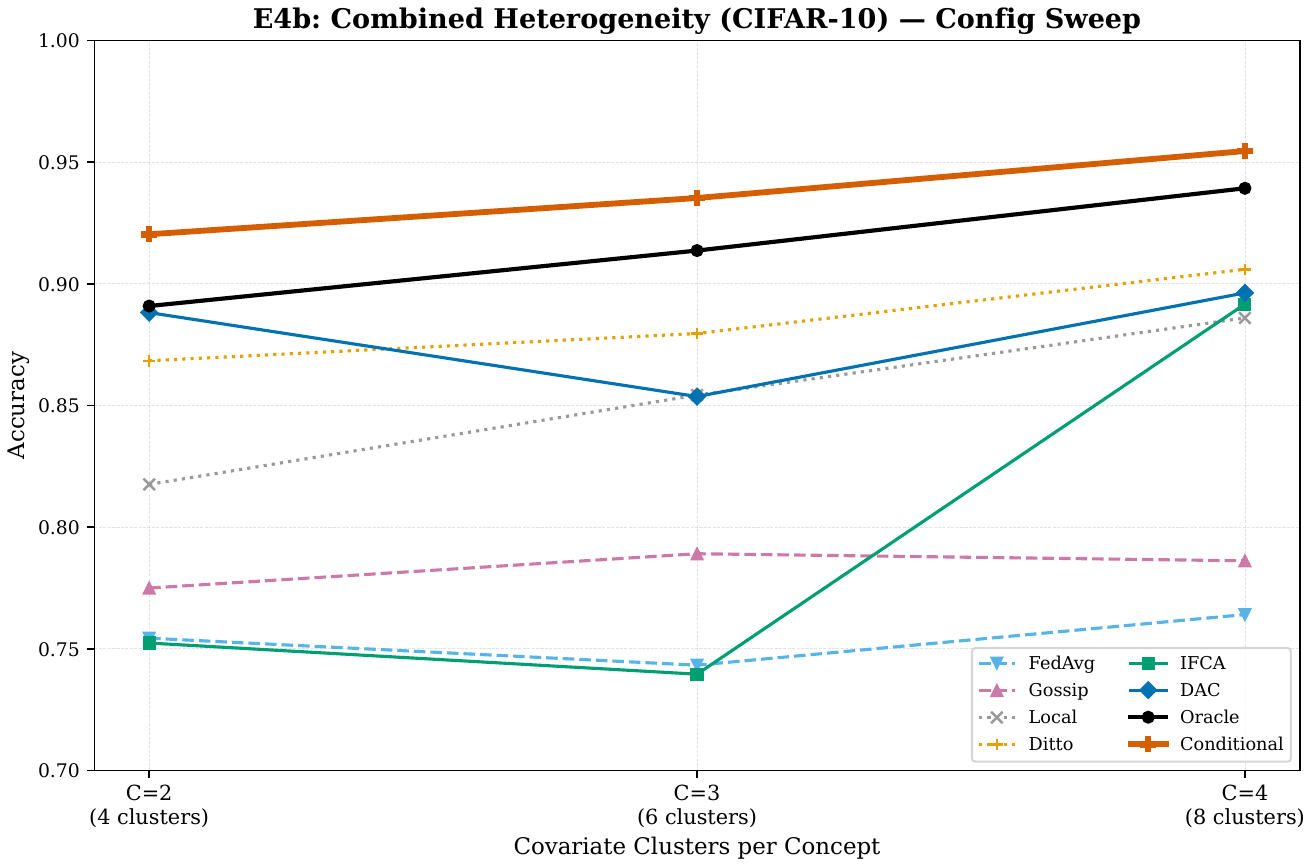}
\caption[E4b: Combined heterogeneity config sweep]{E4b: Combined heterogeneity configuration sweep. Conditional exceeds Oracle across all configurations.}
\label{fig:e4b_config}
\end{figure}

\subsection{E4b: Sparsity Analysis}

\begin{table}[htbp]
\centering
\caption[E4b: Combined heterogeneity sparsity sweep]{E4b: Combined Heterogeneity --- Sparsity Sweep (C=2, CIFAR-10)}
\label{tab:e4b_sparsity}
\begin{tabular}{l|cccc}
\toprule
Method & Rich & Medium & Sparse & Very Sparse \\
& (20 clients) & (40 clients) & (100 clients) & (200 clients) \\
\midrule
Local       & .825 & .793 & .745 & .706 \\
FedAvg      & .753 & .742 & .732 & .727 \\
Gossip      & .770 & .741 & .686 & .647 \\
Ditto       & .869 & .833 & .771 & .737 \\
IFCA        & .753 & .747 & .734 & .725 \\
DAC         & .888 & .846 & .795 & .764 \\
\midrule
\rowcolor{oraclecolor!20}
Oracle      & .900 & .901 & .907 & .895 \\
\rowcolor{bestcolor!15}
Conditional & \best{.919} & \best{.919} & \best{.900} & \best{.856} \\
\bottomrule
\end{tabular}
\end{table}

\textbf{Key observations:}
\begin{itemize}
    \item Conditional beats Oracle at Rich and Medium ($+1.9\%$), and matches at Sparse
    \item IFCA fails completely (ARI=0.0) on this combined structure---the multi-dimensional heterogeneity prevents successful clustering
    \item Ditto achieves reasonable performance (86.9\% Rich) but degrades with sparsity
\end{itemize}

\begin{figure}[htbp]
\centering
\includegraphics[width=0.85\textwidth]{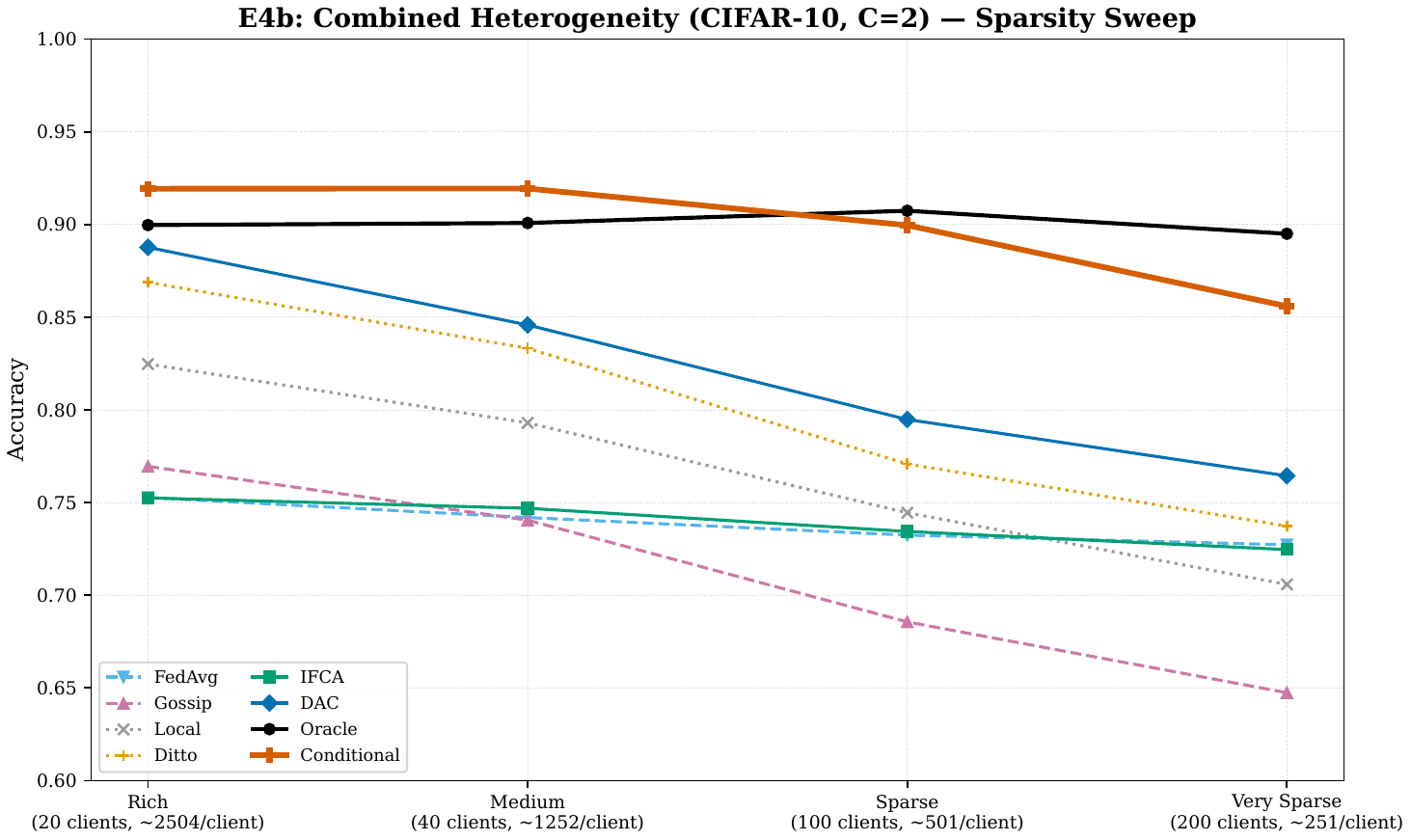}
\caption[E4b: Combined heterogeneity sparsity sweep]{E4b: Combined heterogeneity sparsity sweep (C=2). Conditional beats Oracle at Rich/Medium and remains competitive at sparse settings.}
\label{fig:e4b_sparsity}
\end{figure}

\section{Summary of Results}
\label{sec:summary}

\subsection{Conditional vs Oracle Performance}

\begin{table}[htbp]
\centering
\caption[Conditional vs Oracle summary]{Conditional vs Oracle Performance Across All Experiments}
\label{tab:supp_summary}
\begin{tabular}{l|c|l}
\toprule
Experiment & Conditional vs Oracle & Key Observation \\
\midrule
E1: Label Shift & Matches & Sparsity-invariant \\
E2a: Covariate (Subsampling) & Matches & Robust across K \\
E2b: Covariate (Rotation) & Matches & Works on transformed data \\
E3a: Semantic Concept Shift & Matches & Handles concept ambiguity \\
E3b: Label Permutation & \textbf{Beats} (+1--6\%) & Statistics encode more than cluster ID \\
E4a: Cross-Dataset Domain & Matches & Clear domain separation \\
E4b: Combined Heterogeneity & \textbf{Beats} (+1.5--3\%) & Multi-dimensional information captured \\
\bottomrule
\end{tabular}
\end{table}

\subsection{Method Rankings}

Based on mean accuracy across all experiments:

\begin{enumerate}
    \item \textbf{Conditional} --- Consistently matches or beats Oracle without knowing cluster assignments
    \item \textbf{Oracle} --- Upper bound requiring ground-truth clusters
    \item \textbf{DAC} --- Strong decentralized baseline with soft clustering
    \item \textbf{IFCA} --- Good when clustering succeeds; fails on complex structures
    \item \textbf{Local} --- Benefits from homogeneous local data
    \item \textbf{Ditto} --- Moderate personalization benefit
    \item \textbf{Gossip} --- Collapses under heterogeneity at sparse settings
    \item \textbf{FedAvg} --- Worst under heterogeneity
\end{enumerate}

\subsection{Key Takeaways}

\begin{enumerate}
    \item \textbf{Conditional is the only method that consistently matches Oracle} without requiring cluster information. It achieves this by learning to condition on client-specific PCA statistics.

    \item \textbf{Conditional can beat Oracle} when heterogeneity is multi-dimensional (E3b, E4b), because the statistics capture richer information than discrete cluster membership.

    \item \textbf{Conditional is sparsity-invariant}: Performance remains stable from Rich ($\sim$6000 samples/client) to Super Sparse ($\sim$300 samples/client), while other methods degrade significantly.

    \item \textbf{Clustering methods struggle with complex heterogeneity}: IFCA achieves ARI=1.0 on simple domain shift but ARI=0.0 on combined heterogeneity.

    \item \textbf{FedAvg and Gossip collapse under concept shift}: When label semantics differ across clients, naive averaging destroys information.
\end{enumerate}

\section{Baseline Implementation Details}
\label{sec:baselines}

This section documents the implementation of each baseline method, including deviations from the original papers and their justifications.
Ensuring faithful baseline implementations is critical for fair comparison; we describe the choices made and their impact on results.

\subsection{IFCA (Ghosh et al., NeurIPS 2020)}

We follow the algorithm described in~\cite{ghosh2020ifca}.
In each communication round, each client evaluates the training loss under each of the $K$ cluster models (E-step) and selects the model with the lowest loss for local SGD updates (M-step).
Cluster models are updated by averaging the parameters of all clients assigned to that cluster, weighted by local dataset size.
We use random initialization of cluster models without multiple restarts.

Our implementation captures the core IFCA algorithm behavior: it achieves perfect clustering (ARI${}=1.0$) on experiments with clear cluster structure (E4a domain shift, E3b label permutation with $K \leq 4$), and degrades gracefully when the cluster structure is ambiguous (ARI${}=0.0$ on E4b combined heterogeneity).
The lack of multiple restarts may affect cluster discovery in borderline cases, but we always provide the true number of clusters $K$, giving IFCA an advantage that would be unavailable in practice.

\subsection{Ditto (Li et al., ICML 2021)}

We follow the alternating optimization procedure described in Algorithm~1 of~\cite{li2021ditto}.
In each communication round, each client performs two updates: (1)~a \emph{personal update} that optimizes a local objective regularized toward the current global model, and (2)~a \emph{global update} using standard local SGD.
The personal models are persistent across rounds and the regularization loss is:
\begin{equation}
\mathcal{L}_{\text{personal}} = \mathcal{L}_{\text{task}} + \frac{\lambda}{2} \| w - w_{\text{global}} \|^2
\end{equation}
We use $\lambda = 1.0$ and plain SGD (no momentum), consistent with the original paper, with $n_{\text{personal\_epochs}} = 5$.

\textbf{Note on implementation choice.}
We initially implemented a two-stage variant (full FedAvg convergence followed by personalization fine-tuning with $\lambda = 0.1$), which the original paper explicitly identifies as a weaker baseline (referred to as ``local fine-tuning'' in Section~4 of~\cite{li2021ditto}).
Switching to the paper-faithful alternating approach yielded consistent improvements, particularly on harder tasks: $+$2--10pp on CIFAR-10 label shift, $+$5--10pp on CIFAR-10 sparsity settings, and $+$3--9pp on combined heterogeneity (E4b).
On easier tasks (MNIST), the difference was negligible ($<$0.3pp).
All results reported throughout this paper use the paper-faithful alternating implementation.

\subsection{DAC (Listo Zec et al., FL-IJCAI 2022)}

We use a practical variant of DAC~\cite{zec2022dac} with cosine-similarity-based neighbor discovery in model parameter space, deterministic top-$k$ neighbor selection, and uniform weight averaging.
This differs from the original paper's loss-based similarity evaluation, probabilistic neighbor sampling from learned priors, and softmax-weighted aggregation with temperature.

\textbf{Note on implementation choice.}
We implemented and evaluated both the paper-faithful version (loss-based similarity, probabilistic sampling, softmax-weighted aggregation with $\tau = 10$, $n_{\text{sampled}} = 5$) and the simplified cosine-similarity variant.
The paper-faithful version performed consistently worse across all 97 experimental configurations (mean accuracy drop of 4.2pp), with catastrophic degradation in high-sparsity regimes (up to $-$39pp on E3b MNIST Super Sparse).
We attribute this to the loss-based evaluation requiring each client to evaluate multiple neighbor models on local data, which becomes noisy with limited samples, leading to poor neighbor selection and unstable prior updates.
The cosine-similarity approach operates on full model parameter vectors and does not suffer from this data-dependent noise.
We therefore use the cosine-similarity variant throughout, which provides a stronger baseline for comparison.

\end{document}